\documentclass[journal,onecolumn]{IEEEtran}
\usepackage[utf8]{inputenc} 
\usepackage[T1]{fontenc}    
\usepackage{hyperref}       
\usepackage{url}            
\usepackage{booktabs}       
\usepackage{amsfonts,amsmath,amsthm,amssymb}       
\usepackage{nicefrac}       
\usepackage{microtype}      
\usepackage{lipsum}		
\usepackage{graphicx}
\usepackage{natbib}
\usepackage{doi}
\usepackage[skins]{tcolorbox}
\usepackage{tikz}
\usetikzlibrary{arrows.meta, positioning, shapes.geometric, decorations.pathreplacing, calc, fit, shapes.multipart}
\usepackage{adjustbox,subcaption}
\usepackage{mathtools}

\newtheorem{definition}{Definition}

\newtheorem{theorem}{Theorem}
\newtheorem{lemma}{Lemma}
\newtheorem{proposition}{Proposition}
\newtheorem{corollary}{Corollary}
\newtheorem{principle}{Principle}
\newtheorem{remark}{Remark}
\newtheorem{example}{Example}

\title{Information Topology}

\author{ \href{https://orcid.org/0000-0003-2067-2763}{\includegraphics[scale=0.06]{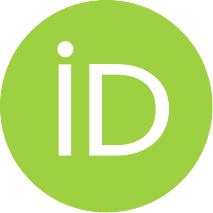}\hspace{1mm}Xin Li}\thanks{This work was partially supported by NSF IIS-2401748 and BCS-2401398. The author has used ChatGPT 5 and Gemini 2.5 Pro models to assist the development of theoretical ideas and visual illustrations presented in this paper.} \\
	Department of Computer Science\\
	University at Albany\\
	Albany, NY 12222 \\
	\texttt{xli48@albany.edu} 
}

\hypersetup{
pdftitle={A template for the arxiv style},
pdfsubject={q-bio.NC, q-bio.QM},
pdfauthor={David S.~Hippocampus, Elias D.~Striatum},
pdfkeywords={First keyword, Second keyword, More},
}

\tikzset{mechanism/.style={rectangle, draw=blue!60, fill=blue!10, thick, minimum height=2em, minimum width=4em, align=center}}
\begin{document}
\maketitle

\begin{abstract}
We introduce \emph{Information Topology}: a framework that unifies information theory and algebraic topology by treating \emph{cycle closure} as the primitive operation of inference. The starting point is the \emph{dot-cycle dichotomy}, which separates pointwise, order-sensitive fluctuations (dots) from order-invariant, predictive structure (cycles). Algebraically, closure is the cancellation of boundaries ($\partial^2=0$), which converts transient histories into stable invariants.
Building on this, we derive the \emph{Structure-Before-Specificity} (SbS) principle: stable information resides in nontrivial homology classes that persist under perturbations, while high-entropy contextual details act as scaffolds. The \emph{Context-Content Uncertainty Principle} (CCUP) quantifies this balance by decomposing uncertainty into contextual spread and content precision, showing why prediction requires invariance for generalization. Measure concentration onto residual invariant manifolds explains \emph{order invariance}: when mass collapses to a narrow tube around a closed cycle, reparameterizations of micro-steps leave predictive functionals unchanged.
We then define \emph{homological capacity}, the topological dual of Shannon capacity, as the sustainable number of independent informational cycles supported by a system. This capacity links dynamical (KS) entropy to structural (homological) capacity and refines Euler characteristics from a ``net'' summary to a ``gross'' count of persistent invariants. Finally, we illustrate the theory across three domains where \emph{more is different}: \textbf{visual binding}, \textbf{working memory}, and \textbf{access consciousness}. Together, these results recast inference, learning, and communication as \emph{topological stabilization}: the formation, closure, and persistence of informational cycles that make prediction robust and scalable.
\end{abstract}

\vspace{\baselineskip}


\section{Introduction}
\label{sec:intro}

Information theory has long provided a quantitative language for uncertainty, communication, and inference \cite{shannon1948mathematical}.
However, its classical formulation, rooted in symbol statistics and channel capacity, remains fundamentally
\emph{pointwise} \cite{cover1999elements}: information is assigned to discrete events or random variables,
and their relations are captured through additive quantities such as entropy or mutual information.
This dot-based view has yielded extraordinary success in data compression \cite{sayood2017introduction}, reliable communication \cite{gallager1968information}, and statistical learning \cite{vapnik2013nature},
but it treats structure as an afterthought.
When information is distributed across loops, recurrences, or higher-order relations,
pointwise measures fail to capture the invariants that persist across transformations.

\noindent\textbf{From dots to cycles.}
Inspired by John Wheeler's {\em It-from-Bit} dictum \cite{wheeler1990information}, we propose to extend Shannon’s theory from the \emph{dot domain} of local statistics
to the \emph{cycle domain} of relational invariants.
A dot represents localized uncertainty, whereas a cycle represents a constraint that
\emph{survives} under boundary operations.
Topologically, this persistence is expressed as the identity $\partial^2 = 0$ \cite{hatcher2002algebraic}:
boundaries of boundaries vanish, allowing cycles to encode closed,
self-consistent relations among informational elements. Algebraically, this forces a dichotomy in how information is 
organized: fragments that do not close into cycles are annihilated, while 
those that do close persist. We call this the \emph{dot-cycle dichotomy}: 
\emph{dots} are the ephemeral remnants of open chains; while \emph{cycles} are closed structures in the kernel of $\partial$. 
In this sense, the topological kernel $\ker \partial$ generalizes the informational kernel
of sufficient statistics \cite{ay2015information}: a cycle carries precisely the amount of information
that remains invariant under reparameterization or observation. 

\noindent\textbf{Cycle closure as the primitive of inference.}
The central idea of \emph{Information Topology} is that inference corresponds to
\textbf{cycle closure}, the cancellation of informational boundaries through mutual constraint.
When two informational chains share a common boundary, closure fuses them into a higher-order invariant.
This operation replaces symbolic concatenation with \emph{homological composition},
turning prediction, communication, and memory into dynamic processes of boundary alignment.
Cycle closure provides a topological interpretation of the
Data Processing Inequality \cite{cover1999elements} and other information-preserving transformations \cite{amari2016information}:
information cannot increase along open chains, but it stabilizes when cycles close. Cognitively, cycles become memory not by mere
storage of instances, but by persisting as structural invariants. They collapse
many distinct inputs into the same homology class, compressing specificity
into structure. This compression is the essence of memory \cite{vecchi2020memory}: it prunes away
ephemeral variation, while retaining the persistent core that can be reused
for future inference.

\noindent\textbf{Structure before specificity.}
Building upon this topological foundation, we articulate the
\emph{Structure-Before-Specificity (SbS)} or \emph{Semanticc-Before-Syntax (SbS)} principle.
Stable information corresponds to low-entropy, persistent structures (nontrivial homology classes),
while transient, high-entropy components serve as exploratory scaffolds that
enable adaptation and differentiation.
SbS formalizes an intuition long implicit in coding theory and learning \cite{tishby2000information}:
efficient representations emerge not by enumerating all details,
but by discovering invariant backbones that survive perturbations. The SbS principle is also consistent with Pinker's semantic bootstrapping hypothesis in language evolution \cite{fitch2010evolution}, which states how semantics emerges first at the group level as shared
cycles of interaction, with syntax arising later as a residual coding overlay \cite{pinker1984semantic}.

\noindent\textbf{Context–Content Uncertainty Principle.}
The interaction between structure and specificity is governed by the
\emph{Context–Content Uncertainty Principle (CCUP)} \cite{li2025CCUP}, which reframes inference as a directional and dynamic alignment problem. Under CCUP, stable and low-entropy content provides the scaffolding necessary to constrain and disambiguate high-entropy contextual input.
Total uncertainty decomposes into contextual and content components,
whose mutual alignment defines the degree of closure.
Where Shannon’s mutual information quantifies statistical dependency \cite{shannon1948mathematical},
CCUP quantifies \emph{topological coherence} between exploratory and persistent informational fields.
It extends the notion of free-energy minimization \cite{friston2006free} to the information-topological domain,
revealing that predictive efficiency arises from the mutual cancellation of contextual and contentual boundaries. SbS and CCUP are not competing principles
but nested layers: SbS provides the developmental ordering, while CCUP
provides the dynamical law.

\noindent\textbf{Homological capacity and information geometry.}
We introduce the notion of \emph{homological capacity}, the topological dual of Shannon capacity, to measure the maximal number of independent informational cycles that can coexist within a system. While Amari’s Information Geometry \cite{amari2016information} provides a local differential structure for statistical manifolds through the Fisher metric and $\alpha$-connections, it remains fundamentally metric-based: information is quantified by how distinguishable nearby probability distributions are. In contrast, homological capacity is topological and global; it quantifies how many independent informational cycles persist after all boundary cancellations. Whereas curvature measures how information bends under smooth transformations, homological capacity measures how information persists under structural deformation. Topologically, homological capacity generalizes the Euler characteristic \cite{hatcher2002algebraic},
providing a structural index of how many dimensions of information can be
stably encoded without destructive interference. Together, Euler characteristic and homological capacity echo the integration–differentiation balance seen in theories of consciousness and cognition (e.g., Integrated Information measure \cite{tononi2016integrated}), but now cast in a topological form directly measurable via persistent homology \cite{edelsbrunner2008persistent}.

\noindent\textbf{Contributions.}
This paper establishes the mathematical and conceptual foundation of Information Topology.
We (1) formalize the dot–cycle dichotomy as a bridge between local entropy and global invariance;
(2) define cycle closure as the generative mechanism of inference;
(3) derive the SbS and CCUP principles as topological laws of information alignment; 
(4) introduce homological capacity as the dual invariant to Shannon capacity; and
(5) apply the information topological framework into cognitive science. 
Together, these results suggest that information is not merely a measure of uncertainty,
but a topological quantity whose stability arises from the closure of cycles in the informational manifold. The rest of this paper is organized as shown in Fig. \ref{fig:teaser}.

\begin{figure}
    \centering
    \resizebox{0.5\linewidth}{!}{
\begin{tikzpicture}[
  >=Latex,
  node distance=8mm,
  block/.style={draw, rounded corners, align=center, inner sep=5pt, font=\small, fill=white, text width=8cm},
  hub/.style={block, fill=blue!5, text width=8cm},
  pillar/.style={block, fill=gray!5, text width=8cm},
  outcome/.style={block, fill=green!10, text width=8cm},
  mechanism/.style={block, fill=yellow!15, text width=8cm}
]


\node[block, fill=purple!8] (dichotomy) {%
\textbf{Invariance via Broken Symmetry (Sec.~\ref{sec:2})}\\
Dot-Cycle Dichotomy \\
Dots: local statistics (entropy of symbols) \quad vs.\quad
Cycles: relational invariants in $\ker \partial$\\
$C_k(\mathcal{Z}),~\partial,~\ker\partial,~\mathrm{im}\,\partial,\; H_k=\ker\partial/\mathrm{im}\,\partial$
};

\node[mechanism, fill=orange!8, below=8mm of dichotomy] (closure) {%
\textbf{Cycle Closure \& Boundary Cancellation (Sec.~\ref{sec:3})}\\
$\partial^2=0$ induces order invariance; boundaries of boundaries vanish\\
Inference as boundary alignment $\Rightarrow$ closed informational loops
};

\node[pillar, below=8mm of closure] (sbs) {%
\textbf{Structure-Before-Specificity (Sec.~\ref{sec:SbS})}\\
Persistent structures: low-entropy, nontrivial $[\gamma]\in H_k$\\
Transient scaffolds: high-entropy exploration $\to$ refinement of structure\\
equivalent to Semantics-before-Syntax
};

\node[pillar, below=8mm of sbs] (ccup) {%
\textbf{Context-Content Uncertainty Principle (Sec.~\ref{sec:ccup})}\\
Synchronization as context filter and recurrence as content filter \\
Decomposition of uncertainty; alignment $\Leftrightarrow$ cycle coherence\\
Boundary cancellation between context ($\Psi$) and content ($\Phi$)
};

\node[hub, below=10mm of ccup] (cycles) {%
\textbf{Cycle-Based Memory (SbS\,\&\,CCUP Sec.~\ref{sec:cycle-memory})}\\
Alignment checkpoints (CCUP) $+$ reuse via closure $\Rightarrow$ persistence\\
Topological stabilization of information across paths and scales
};

\node[block, fill=cyan!10, below=8mm of cycles] (capacity) {%
\textbf{Homological Capacity (Sec.~\ref{sec:5})}\\
$C_H$ = rank of independent persistent cycles (dual to Shannon capacity)\\
Euler duality: structural index via $\chi$ balances exploration vs.\ persistence
};

\node[outcome, below=8mm of capacity] (out) {%
\textbf{Applications \& Outlook (Sec.~\ref{sec:6})}\\
Perception, action, learning, and planning as topological stabilization\\
Cycle-based memory explains binding, working memory, and access consciousness
};

\draw[->, thick] (dichotomy.south) -- (closure.north);
\draw[->, thick] (closure.south) -- (sbs.north);
\draw[->, thick] (sbs.south) -- (ccup.north);
\draw[->, thick] (ccup.south) -- (cycles.north);
\draw[->, thick] (cycles.south) -- (capacity.north);
\draw[->, thick] (capacity.south) -- (out.north);

\end{tikzpicture}
}
 \caption{Information Topology: an end-to-end view aligned with the paper’s sections.
Broken symmetry leads to residual invariance; prediction requires order invariance for generalization. Dots (local statistics) and cycles (global invariants) are linked by cycle closure
($\partial^2\!=\!0$), which stabilizes information. SbS provides persistent structure,
CCUP governs alignment between context and content, and \emph{homological capacity} quantifies
how many independent informational cycles a system can stably support, yielding robust prediction and planning.}
    \label{fig:teaser}
\end{figure}

\section{Residual Invariance from Broken Symmetry}
\label{sec:2}

\subsection{Dot-Cycle Dichotomy}
\label{sec:2a}

\noindent\textbf{From \emph{It-from-Bit} to \emph{Cycle-from-Boundary}.}
Wheeler’s dictum “it-from-bit” \cite{wheeler1990information} proposes that physical reality arises from binary distinctions.
In information topology, the analog of a binary distinction is the
\emph{boundary operation} $\partial$.
Its algebraic consistency, $\partial^2=0$, ensures that
boundaries of boundaries vanish, providing a structural law of
self-consistency.
This closure condition generates a natural hierarchy, which we call the dot-cycle dichotomy (Fig. \ref{fig:trivial-vs-nontrivial}):
\emph{dots} (0-chains) represent discrete distinctions, while
\emph{cycles} (elements of $\ker\partial$) represent relations that
remain invariant under boundary reduction.
The transition from bits to cycles marks the passage
from discrete separation to topological coherence \cite{hatcher2002algebraic}.

\noindent\textbf{Set-up: State Space, Synchronization, and Recurrence.}
Let the state of a computational or physical system be represented by a point in a
metric space $(\mathcal{Z}, d)$.
A computation unfolds as a time-ordered sequence of states
$S = \{z_1, z_2, \dots, z_N\}$ with $z_i \in \mathcal{Z}$.
Two fundamental filters act on this sequence:
1)~\textbf{Synchronization:}
A gating mechanism that selects temporally coincident or phase-aligned events.
Formally, define a coincidence function
$\phi(z_i, z_j)\!\to\!\{0,1\}$ that is non-zero only if the two states
occur within a time window $\Delta t$ and satisfy a phase-alignment criterion
(e.g., spike coincidence or oscillatory coherence
\cite{abeles1982role,konig1996integrator}).
Synchronization selects events that happen \emph{together}.
2)~\textbf{Recurrence:}
A condition that selects states revisiting a region of $\mathcal{Z}$.
The sequence $S$ exhibits recurrence if there exist $i\!\ll\!j$
with $d(z_i, z_j)<\epsilon$ for some small $\epsilon$.
Recurrence selects events that \emph{return} \cite{walters1982introduction}.
Together, synchronization and recurrence identify subsets of the state space
that are both temporally and spatially coherent, forming the raw substrate
for topological invariants.

\noindent\textbf{Dot–Cycle Dichotomy.}
The synchronized-recurrent subsets can be embedded in a simplicial complex
$\mathcal{K}$ built from the data \cite{hatcher2002algebraic}.
In a Vietoris–Rips complex, the states $\{z_i\}$ are vertices, and edges
connect any pair with $d(z_i, z_j)<\epsilon$.
Within $\mathcal{K}$, a \emph{$k$-cycle} is a $k$-dimensional chain
$c$ satisfying $\partial c = 0$.
Intuitively, a 1-cycle is a closed loop, a 2-cycle a closed surface, and so on.
Cycles that are not exact ($c\!\notin\!\mathrm{im}\,\partial$) form the nontrivial homology classes
$H_k(\mathcal{K})=\ker\partial_k/\mathrm{im}\,\partial_{k+1}$
that quantify persistent $k$-dimensional ``holes'' in the state space
\cite{edelsbrunner2008persistent}.
These classes constitute the stable, invariant structures extracted by
the computation.
Individual states or dots are transient, order-dependent, and
high-entropic: they correspond to the \emph{contextual} fluctuations of experience.
Closed cycles, by contrast, are order-invariant, low-entropic, and
structurally persistent: they represent the stable \emph{contents}
that survive perturbations.
Synchronization produces simultaneity and recurrence produces closure; together they transform fleeting dots into persistent cycles, which form the basis of memory \cite{vecchi2020memory}.

\begin{principle}[Dot–Cycle Dichotomy as the Basis of Memory]
\label{prin:dot-cycle}
A system’s information decomposes into transient high-entropy
\emph{contexts} (dots) and persistent low-entropy
\emph{contents} (cycles).
Computation by cycle closure filters out trivial fluctuations,
retaining only closed, nontrivial homology classes that form the
substrate of memory.
Prediction arises by advancing along these invariant classes,
while learning corresponds to the formation of new ones.
\end{principle}

\begin{figure}[t]
\centering
\resizebox{\linewidth}{!}{
\begin{tikzpicture}[
  scale=1.0,
  every node/.style={font=\small},
  panel/.style={rounded corners=6pt, draw=gray!60, line width=0.8pt, minimum width=6.2cm, minimum height=4.6cm},
  cycletriv/.style={line width=1.4pt, blue!70},
  cyclenontriv/.style={line width=1.4pt, red!70},
  fillface/.style={fill=blue!12, draw=none},
  holedisc/.style={fill=white, draw=gray!65, line width=0.9pt}
]

\begin{scope}[shift={(-4.,0)}]
  \node[panel] (LeftBox) at (0,0) {};
  \node[gray!70] at (0,2.5) {\bfseries Trivial 1-cycle (null-homologous)};

  \fill[fillface] (0,0) circle (1.3);

  \draw[cycletriv] (0,0) circle (1.3);

  \node at (0,0) {$S$};
  \node[blue!70] at (0,-1.8) {$\gamma=\partial S,\quad [\gamma]=0 \ \text{in}\ H_1$};

  \draw[->, gray!70] (1.4,0.2) -- (0.7,0.1);
  \draw[->, gray!70] (-1.4,-0.1) -- (-0.7,-0.05);
\end{scope}

\begin{scope}[shift={(4.,0)}]
  \node[panel] (RightBox) at (0,0) {};
  \node[gray!70] at (0,2.5) {\bfseries Nontrivial 1-cycle (not a boundary)};

  \fill[gray!06] (-2.8,-2.0) rectangle (2.8,2.0);

  \node[holedisc] (Hole) at (0,0) [circle, minimum width=2.0cm, minimum height=2.0cm] {};

  \draw[cyclenontriv] (0,0) circle (1.4);

  \node[gray!70] at (0,0) {\scriptsize hole};
  \node[red!70] at (0,-1.8) {$[\gamma]\neq 0 \ \text{in}\ H_1$};
\end{scope}

\end{tikzpicture}
}
\caption{\textbf{Dot-Cycle Dichotomy.} 
\emph{Left:} A cycle that is the boundary of a filled region ($\gamma=\partial S$) is 
\emph{null-homologous} and therefore trivial in $H_1$: it can be “canceled” as a boundary when becoming a dot. 
\emph{Right:} A cycle encircling a hole is not the boundary of any 2-chain in the space, 
so it represents a \emph{nontrivial} class in $H_1$. In our framework, trivial cycles 
correspond to high-entropy, short-lived scaffolds ($\Psi$) that collapse under boundary 
cancellation ($\partial^2=0$), whereas nontrivial cycles correspond to low-entropy content 
invariants ($\Phi$) that persist as memory.}
\vspace{-0.2in}
\label{fig:trivial-vs-nontrivial}
\end{figure}

\noindent\textbf{From Structural Invariance to Dynamical Emergence.}
Having defined cycles as the low-entropy carriers of structural information, we now turn to their dynamical selection.
In practice, invariant cycles do not preexist as abstract objects; they
\emph{emerge} when continuous symmetries in the system’s dynamics are
broken \cite{goldstone1962broken}. 
Symmetry breaking confines trajectories to low-dimensional manifolds,
where high-entropy contextual variability collapses and residual invariants
persist as closed recurrent loops, attractors, or conserved quantities \cite{fenichel1971persistence}.
These surviving structures instantiate the cycle classes introduced in
Sec.~\ref{sec:2a}, converting topological coherence
($\partial^2=0$) into a measurable concentration of probability mass.
We next formalize this process using the notions of non-ergodicity,
context-dependent potentials, and entropy–concentration tradeoffs.

\subsection{Non-Ergodicity, Context, and Concentration}
\label{sec:2b}

We retain the notation of Sec.~\ref{sec:2a}.
Let $\Psi\in\mathcal{C}$ be a slowly varying \emph{context} that parametrizes the transition kernel $\mathsf{P}_\Psi$ on $(\mathcal{Z},d)$.
Under symmetry breaking, the dynamics are typically \emph{non-ergodic}: for fixed $\Psi$ there exist finitely many invariant components $\{\mathcal{A}_j(\Psi)\}_{j=1}^J$ with ergodic invariant measures $\{\nu_j^\Psi\}$ whose supports contain recurrent \emph{cycle backbones} $\gamma_j\subset\mathcal{Z}$ (for clarity, take $\gamma_j$ 1D).
The stationary law then decomposes as a context-weighted mixture
$\mu_\Psi \;=\; \sum_{j=1}^J \pi_j(\Psi)\,\nu_j^\Psi,
\sum_{j=1}^J \pi_j(\Psi)=1$,
where $\pi_j(\Psi)$ encodes the selection bias induced by $\Psi$.
Symmetry breaking both \emph{selects} (via $\pi_j$) and \emph{stabilizes} (via $\nu_j^\Psi$) residual invariants, preparing the ground for concentration of mass onto the cycle supports $\{\gamma_j\}$.
We formalize this concentration next.

The collection of \emph{cycle supports}
$\mathcal{C}_\Psi=\{\gamma_j\}$ will play the role of residual structural
backbones of the dynamics.  Intuitively, broken symmetry within a given context
selects and stabilizes a subset of these cycles, producing distinct,
context-conditioned modes of recurrence.
The key insight is that {\em when global symmetry is broken, dynamics do not become random, but instead collapse onto low-dimensional invariant manifolds that carry the remaining order} \cite{jain2022compute}.
We formalize this notion by describing how probability mass
concentrates around cycle supports as entropy decreases, which leads to the following definition.

\begin{definition}[Concentration on cycles]
For $\epsilon>0$, define the tubular neighborhood
$N_\epsilon(\gamma)=\{z:d(z,\gamma)\le \epsilon\}$.
We say that $\mu_\Psi$ \emph{concentrates} on $\gamma$ at rate $\kappa(\epsilon)$
if
$\mu_\Psi(N_\epsilon(\gamma))\ge 1-\exp(-\kappa(\epsilon)),
\text{with } \kappa(\epsilon)\to\infty \text{ as } \epsilon\downarrow 0$.
\end{definition}

\noindent
Concentration quantifies the degree to which uncertainty is restricted to a narrow
tube surrounding an invariant cycle.
The following lemma connects symmetry breaking with this concentration behavior.

\begin{lemma}[Broken symmetry $\Rightarrow$ entropy reduction $\Rightarrow$ concentration]
\label{lem:entropy-concentration}
Suppose that context $\Psi$ induces a potential
$U_\Psi:\mathcal{Z}\to\mathbb{R}$ giving rise to the Gibbs measure
$\mu_{\Psi,\beta}(\mathrm{d}z)\propto \exp(-\beta U_\Psi(z))\,\mathrm{d}z$
with $\beta>0$ quantifying the strength of symmetry breaking
(i.e., inverse temperature).
If $U_\Psi$ attains its minima along a cycle $\gamma$
and $\nabla^2 U_\Psi$ is positive definite in directions normal to $\gamma$, then we have:
1) The entropy
$H(\mu_{\Psi,\beta})
= \text{const}-\mathbb{E}_{\mu_{\Psi,\beta}}[\log \mu_{\Psi,\beta}]$
is strictly decreasing in $\beta$;  stronger symmetry breaking reduces entropy.
2) There exist constants $c,c'>0$ such that for every Lipschitz $f$,
$\mathbb{P}_{\mu_{\Psi,\beta}}\big(|f-\mathbb{E}f|\ge r\big)
\le 2\,\exp(-c\,\beta\,r^2),
\mu_{\Psi,\beta}(N_\epsilon(\gamma))
\ge 1- \exp(-c'\beta\,\epsilon^2)$.
The probability mass concentrates around $\gamma$
at a sub-Gaussian rate controlled by $\beta$.
\end{lemma}

\noindent
The lemma formalizes an entropy–concentration tradeoff \cite{ledoux2001concentration}: as symmetry breaking increases, entropy decreases,
and the distribution contracts around a subset of
low-dimensional invariants.
This marks the transition from diffuse exploration to structured persistence.

\begin{principle}[Broken Symmetry and Residual Invariants]
\label{prin:broken-symmetry}
When high-dimensional dynamics are subjected to symmetry breaking,
variability collapses yet \emph{residual invariants} remain.
These invariants typically manifest as closed loops, attractors,
or conserved quantities that survive the collapse of degrees of freedom.
They serve as \emph{cycle supports} $\{\gamma_j\}$ along which
probability mass concentrates as entropy decreases.
In this regime, order-specific fluctuations are suppressed,
and the invariant cycle class $[\gamma]$ becomes a sufficient carrier
of predictive information. Therefore, broken symmetry constitutes the gateway from transient dynamics to stable, predictively sufficient invariants, the essential step
from ergodic randomness to structured memory.
\end{principle}

\noindent
Principle~\ref{prin:broken-symmetry} establishes the conceptual link between
symmetry breaking and the emergence of residual invariants that act as
stable carriers of predictive information.
We now turn to a quantitative formulation of this idea.
Once entropy reduction has confined probability mass to the vicinity of
these invariant cycles, fluctuations within the tube become redundant for
prediction, and the system’s effective state space collapses onto a small
set of macroscopic invariants \cite{kaszas2025globalizing}.
The following subsection formalizes this transition from geometric
stability to informational sufficiency, showing how residual invariance
yields topological compression and predictive generalization.

\subsection{Predictive Sufficiency from Residual Invariance}
\label{sec:2c}

The concentration results of Sec.~\ref{sec:2b} imply that once symmetry breaking
has confined probability mass to a narrow tube surrounding a cycle, all local fluctuations within that tube become statistically redundant.
Prediction therefore depends only on the \emph{cycle class} $[\gamma]$,
not on the detailed order of microstates along the trajectory.
The following theorem formalizes this intuition.

\begin{theorem}[Maximal predictive sufficiency of cycle class under concentration]
\label{thm:sufficient-cycle}
Let $Y$ be the task-relevant future variable (e.g., an object parameter or goal outcome),
and assume \emph{conditional stability}:
$p(Y| Z_{0:t},\Psi)=p(Y| [\gamma],\Psi)$ whenever
$\mathrm{dist}(Z_{0:t},\gamma)\le\epsilon$.
If $\mu_{\Psi,\beta}$ concentrates on $\gamma$ at rate $\kappa(\epsilon)$
and $f$ is $1$-Lipschitz in $Z_{0:t}$,
then for any predictor $h$ there exists an invariant predictor $\tilde h$
depending only on the cycle class $[\gamma]$ such that
$\mathbb{E}\,\ell\!\big(Y,\tilde h([\gamma],\Psi)\big)
\;\le\;
\mathbb{E}\,\ell\!\big(Y,h(Z_{0:t},\Psi)\big)
\;+\; O\!\big(e^{-\kappa(\epsilon)}\big)$
for any bounded proper loss $\ell$.
As $\beta\!\to\!\infty$ (stronger symmetry breaking),
$\epsilon\!\downarrow\!0$, and $\kappa(\epsilon)\!\to\!\infty$,
the excess risk vanishes.
\end{theorem}

\begin{remark}[Topological compression meets predictive generalization]
When probability mass collapses to a narrow tube $N_\epsilon(\gamma)$, the fine-grained
order of microstates along the tube ceases to matter for prediction: any two histories
$Z_{0:t},Z'_{0:t}\in N_\epsilon(\gamma)$ differ only by local reparameterizations. Let $G$ denote the group of such reparameterizations
(e.g., permutations that preserve adjacency along $\gamma$). Write
$q: Z_{0:t} \mapsto [\gamma]$ for the projection that forgets order-specific details and
keeps only the \emph{cycle class}.
Under concentration, one has the approximate invariance
$p\big(Y \,\big|\, Z_{0:t},\Psi\big)
\approx
p\big(Y \,\big|\, g\!\cdot\! Z_{0:t},\Psi\big)
\approx
p\big(Y \,\big|\, [\gamma],\Psi\big)
~\text{for all } g\in G$,
so $[\gamma]$ functions as an \emph{approximately sufficient statistic}:
$p(Y | Z_{0:t},\Psi) \approx p\!\big(Y | q(Z_{0:t}),\Psi\big),
I\big(Y;Z_{0:t}\,\big|\,\Psi\big)
\approx
I\big(Y;[\gamma]\,\big|\,\Psi\big)$.
Equivalently, predictive information is \emph{compressed topologically}:
many micro-histories collapse to a single macroscopic invariant $[\gamma]$.  
\end{remark}

\noindent\textbf{Prediction requires order invariance for generalization}.
The topological compression described above concretizes a general
information-theoretic law: successful prediction depends on invariance.
Only when many microscopic trajectories $Z_{0:t}$ collapse onto the
same macroscopic invariant $[\gamma]$ can predictions generalize beyond
particular histories. In this sense, measure concentration performs the
geometric counterpart of the information bottleneck \cite{tishby2000information}:
entropy reduction enforces invariance, and invariance enables prediction.
The following result formalizes this intuition by showing how symmetry
breaking induces \emph{dimension-robust concentration} of a Gibbs measure
around its invariant submanifold.  As the inverse-temperature parameter
$\beta$ increases, high-entropy fluctuations orthogonal to the manifold
are suppressed, confining probability mass to the low-dimensional cycle
that carries the residual invariants.  This concentration mechanism
mathematically realizes the invariance principle stated above.

\begin{theorem}[Dimension-robust concentration on a cycle]
\label{thm:tube}
Let $\mu_\beta(\mathrm{d}x)\propto e^{-\beta U(x)}\mathrm{d}x$ on a Riemannian manifold
$(X,g)$ of dimension $n$, with $U\in C^2$ attaining its minima on a compact $k$-dimensional
submanifold $M$ and $\nabla^2U|_{N_xM}\succeq \lambda_\perp I$ for all $x\in M$.
Then there exist constants $c,c'>0$ independent of $n$ such that for all $\epsilon>0$:
$\mu_\beta\!\big(N_\epsilon(M)\big)\;\ge\; 1-\exp\!\big(-c\,\beta\,\lambda_\perp\,\epsilon^2\big),
\mathbb{P}_{\mu_\beta}\!\left(|f-\mathbb{E}f|\ge r\right)\;\le\;2\exp\!\big(-c'\,\beta\,\lambda_\perp\,r^2\big)$
for every $1$-Lipschitz $f:X\to\mathbb{R}$. Moreover, the marginal of $\mu_\beta$ on $M$
converges (as $\beta\to\infty$) to the Gibbs measure induced by the tangential restriction of $U$;
if $U$ is constant along $M$, this marginal tends to the normalized volume on $M$.
\end{theorem}

\noindent
Theorem~\ref{thm:tube} formalizes the entropy–concentration tradeoff \cite{ledoux2001concentration}:
as symmetry breaking intensifies (larger $\beta$), probability mass
contracts around a low-dimensional manifold $M$ that carries the residual
invariants of the dynamics.  Within this tube, fluctuations orthogonal
to $M$ vanish at a sub-Gaussian rate, ensuring that almost all trajectories
remain confined to a narrow neighborhood of recurrent paths.
This geometric confinement provides the structural substrate for
predictive invariance, but it does not yet explain \emph{why} predictions
become order-independent along these paths.
To understand that invariance, we must ascend from geometry to topology \cite{edelsbrunner2010computational}:
once trajectories close into cycles, the information they carry becomes
governed by the boundary identity $\partial^2=0$,
which enforces consistency of information flow under all local re-orderings.
The next section develops this idea as the principle of
\emph{cycle closure}.

\section{Order Invariance via Cycle Closure}
\label{sec:3}

\subsection{From Concentration to Closure}
\label{sec:3a}

The concentration of probability mass onto residual invariant cycles
(Sec.~\ref{sec:2c}) ensures that most trajectories remain
confined to narrow tubes surrounding closed paths.
However, invariance comes in different forms: 
translation invariance in perception \cite{dicarlo2012does}, scale invariance in physics \cite{wilson1974renormalization}, 
permutation invariance in combinatorial learning \cite{tang2021sensory}. Why does intelligence require specifically \emph{order invariance}? 
The reason lies in the temporal nature of prediction. Observations arrive 
as ordered streams, but their predictive content cannot depend on the 
incidental sequence in which fragments are encountered. A predictor that is 
sensitive to local reordering would constantly change its state with each 
permutation, leading to instability \cite{lee2019set}. For predictive substrates to be 
well-defined, they must remain invariant under reparameterizations, warps, 
and permutations of input order that do not alter the underlying structure \cite{cohen2020regularizing}.

The above observation raises a deeper question: \emph{Why do such recurrent paths support stable, order-independent prediction?}
The answer lies in the principle of \emph{cycle closure}, expressed by
the topological identity
$\partial^2 = 0$,
which states that boundaries of boundaries vanish \cite{wheeler1990information}.
When applied to informational flow, this condition guarantees that
any local ordering of micro-events along a closed trajectory
leaves the global informational invariant unchanged.
Once closure is achieved,
prediction depends only on the \emph{cycle class} $[\gamma]$, not on
the temporal sequence of states that instantiate it. Formally, we have

\begin{proposition}[Cycle formation as the mechanism behind concentration-based invariance]
\label{prop:cycle-concentration}
Assume the setting of Theorem~\ref{thm:tube}, with $M$ a closed $1$-cycle (topological $S^1$)
supporting the invariant set of the dynamics. Let $G$ act by re-ordering micro-steps in the ambient space.
For any $1$-Lipschitz functional $F$ that is constant along reparameterizations of $M$ (goal-centric),
$\sup_{g\in G}\left|\,\mathbb{E}_{\mu_\beta}[F]-\mathbb{E}_{g\cdot \mu_\beta}[F]\,\right|
\;\le\; C\,e^{-c\,\beta}$
and, for any two orderings of the same micro-steps that remain within $N_\epsilon(M)$,
their evaluations differ by at most $O\!\big(\sqrt{\tfrac{1}{\beta\lambda_\perp}}\big)$.
It follows that \emph{order invariance} emerges from high-dimensional concentration onto the cycle.
\end{proposition}

\noindent
Proposition~\ref{prop:cycle-concentration} shows that once probability
mass is concentrated around a closed cycle, the expected value of any
order-invariant functional becomes effectively insensitive to the
specific temporal arrangement of microstates \cite{ledoux2001concentration}.  In other words,
symmetry breaking and subsequent concentration suppress the influence of
permutation noise, enforcing consistency across equivalent orderings.
The following corollary abstracts this result to any invariant functional
$f$ defined on the trajectory space, establishing \emph{functional
order invariance} as a generic consequence of concentration on cycles.

\begin{corollary}[Order invariance as a consequence of concentration]
\label{cor:order-from-concentration}
Let a group $G$ act on trajectories by reordering or reparameterizing
micro-steps.
If $\mu_{\Psi,\beta}$ concentrates on a closed cycle $\gamma$
and $f$ is invariant modulo reparameterization along $\gamma$
(e.g., a goal- or object-centric functional),
then for any $g\in G$,
$\Big|\mathbb{E}_{\mu_{\Psi,\beta}}[f]
- \mathbb{E}_{g\cdot \mu_{\Psi,\beta}}[f]\Big|
\;\le\;O\!\big(e^{-\kappa(\epsilon)}\big)$.
Cycle concentration naturally induces
\emph{functional invariance} to ordering noise and
local temporal perturbations.
\end{corollary}

\noindent\textbf{Interpretation.}
Proposition~\ref{prop:cycle-concentration} quantified order invariance
through geometric curvature and concentration rate, while
Corollary~\ref{cor:order-from-concentration} restated it abstractly:
once probability mass collapses onto a closed cycle, the expectations of
reparameterization-invariant functionals become indistinguishable across
all temporal permutations.  In effect, concentration annihilates order
information, microstate ordering no longer affects prediction. To make this invariance \emph{structural} rather than merely
measure-theoretic, we now recast trajectories in the language of chains \cite{hatcher2002algebraic},
where “dependence on order’’ appears only through a boundary term.
Let $\gamma$ be a recurrent orbit supporting concentrated probability
$\mu_\Psi$.  A finite trajectory segment can be represented as a
1-chain
$c = \sum_i w_i [z_i,z_{i+1}]$,
whose coefficients $w_i$ encode transition weights and whose boundary
$\partial c = \sum_i w_i (z_{i+1} - z_i)$ measures the net imbalance of
successive steps.  For an open chain $\partial c \ne 0$, the associated
informational quantity
$I(c)=\mathbb{E}_{\mu_\Psi}\!\left[\log p(z_{i+1}\!|\!z_i)\right]$
depends explicitly on the ordering of transitions.
Under symmetry breaking, however, concentration around $\gamma$
drives $\partial c \!\to\! 0$, rendering the chain
\emph{closed} in the homological sense.
Once closed, the informational flow through the chain becomes
cyclically consistent:
permuting the order of steps along the loop leaves $I(c)$ invariant.
Closure transforms temporal coherence into structural invariance as a consequence of the Abelian property of the homology group \cite{edelsbrunner2008persistent}.

\begin{theorem}[Cycle Closure Implies Order Invariance]
\label{thm:closure-order}
Let $f$ be a functional on trajectories $f(z_{1:N})$
whose dependence on temporal order enters only through the boundary
$\partial c$ of the associated chain.
If $\partial^2 = 0$ and the trajectory is closed ($\partial c = 0$),
then for any permutation $\pi$ preserving adjacency within the cycle,
$f(z_{1:N}) = f(z_{\pi(1:N)})$.
In other words, closure ensures invariance under temporal
reparameterization: once a trajectory forms a closed cycle, its global
informational content remains unchanged by local rearrangements.
\end{theorem}

\begin{remark}[Topological realization of data processing inequality]
Cycle closure acts as a topological realization of the
\emph{Data Processing Inequality} \cite{cover1999elements}: boundary maps act as information-reducing morphisms,
and the condition $\partial^2=0$ expresses the equality case where
no further loss occurs. Once closure is achieved, the informational
flow becomes conservative, neither amplifying nor diminishing,
and the system attains order invariance.
Information cannot increase along open chains,
but it stabilizes once boundaries cancel.
Therefore, the transition from open to closed chains marks
the boundary between transient processing and persistent memory.
A closed informational loop maintains predictive sufficiency
because every state in the loop implicitly encodes its neighbors.    
\end{remark}

\noindent
Theorem~\ref{thm:closure-order} formalizes order invariance
as a topological property: once trajectories form closed cycles,
the functional outcome becomes independent of local permutations.
This notion admits a precise geometric analog in the language
of sheaves and index theory \cite{grothendieck1958chern}, where closure corresponds to pushforward
and invariance arises from equivariant functoriality.

\begin{proposition}[GRR-style order invariance]
\label{prop:grr-order}
Let $X$ parameterize ordered micro-episodes (glimpses or primitives),
with the symmetric group $S_k$ acting by permutation,
and let $f:X\!\to\!Y$ forget order and map to object or goal space.
For a coherent sheaf $E$ on $X$ (representing evidence or plan),
let $E^{S_k}$ denote the $S_k$-invariant subsheaf.
Then, by the equivariant Grothendieck–Riemann–Roch (GRR) theorem,
$\mathrm{ch}\!\big(f_! E^{S_k}\big)=
f_*\!\Big(\mathrm{ch}(E)^{S_k}\!\cdot\!\mathrm{Td}(T_f)\Big).$
Since the Chern character $\mathrm{ch}$ is multiplicative and commutative,
the right-hand side depends only on the \emph{multiset} of local factors,
not on their order.
Therefore, any global invariant computed via $f_!$, such as an index,
integrated plan, or decision functional, is order-invariant
up to the geometric Todd correction.
\end{proposition}

\noindent
This GRR-style formulation provides an algebraic–geometric mirror
to the topological result of Theorem~\ref{thm:closure-order}:
the invariance of global structure under reordering
is a general property of functorial pushforward \cite{hartshorne2013algebraic}.
Cycle closure in homology parallels order invariance in sheaf cohomology, both expressing that composition and summation commute
when the underlying structure is closed.
Having established that concentration on closed trajectories suppresses order information and that
closure enforces invariance under temporal reparameterization, we now pass from trajectories to
their \emph{structural} representation. The chain–boundary formalism makes this passage explicit:
dependence on ordering enters only through a boundary term, while closure projects dynamics onto
invariant classes. This motivates a homological view in which the informational content of a closed trajectory is represented by its equivalence class $[c]\in H_\bullet(\mathcal{Z})$,
where $H_\bullet(\mathcal{Z})=\ker\partial_\bullet/\mathrm{im},\partial_{\bullet+1}$ encodes invariant cycles modulo trivial boundaries.
In this sense, $H_\bullet$ formalizes the topology of informational persistence, what remains invariant when temporal or contextual order is deformed.
We formalize this intuition next.


\begin{figure*}[h]
\centering
\resizebox{\textwidth}{!}{
\begin{tikzpicture}[
  >=Latex,
  node distance=6mm and 10mm,
  title/.style={font=\small\bfseries, align=center},
  box/.style={draw, rounded corners, align=center, inner sep=5pt, font=\small, fill=white, text width=5.0cm},
  panel/.style={draw, dashed, rounded corners, inner sep=6pt, opacity=0.7},
  lbl/.style={font=\footnotesize\itshape, align=center},
  dot/.style={circle, fill=black, inner sep=0.9pt},
  traj/.style={line width=0.7pt, -{Latex}},
  tube/.style={draw, fill=blue!6, line width=0.6pt},
  loop/.style={draw=blue!60, line width=1.0pt},
  perm/.style={draw, line width=0.7pt, -{Latex}},
  map/.style={draw, thick, -{Latex}}
]

\begin{scope}[shift={(0,-0.05)}]
\node[title] (Atitle) at (0,3.3) {(A) Symmetry breaking};
\node[box, below=3mm of Atitle] (Atext) {Potential $U_\Psi$ breaks symmetry and funnels probability toward a low-dimensional invariant set};

\begin{scope}[yshift=-7mm]
  \draw[fill=gray!10, draw=gray!40] (-2.0,-0.5) .. controls (-1.2,0.6) and (-0.8,0.6) .. (0,0.0)
                                     .. controls (0.8,0.6) and (1.2,0.6) .. (2.0,-0.5) -- cycle;
  \draw[fill=gray!10, draw=gray!40] (-1.6,-0.5) .. controls (-1.0,0.2) and (-0.7,0.2) .. (0,0.0)
                                     .. controls (0.7,0.2) and (1.0,0.2) .. (1.6,-0.5) -- cycle;
  \foreach \x in {-1.4,-0.7,0,0.7,1.4} {
    \draw[traj] (\x,1.0) -- ++(0,-0.7);
  }
\end{scope}

\node[panel, fit=(Atitle)(Atext)] (Abox) {};
\end{scope}

\begin{scope}[shift={(6.6,0)}]
\node[title] (Btitle) at (0,3.3) {(B) Concentration on a cycle};
\node[box, below=3mm of Btitle] (Btext) {Tube $N_\epsilon(\gamma)$ concentrates mass; random perturbations or orthogonal fluctuations are suppressed};

\begin{scope}[yshift=-7mm]
  \draw[tube] (0,0) circle [radius=1.30];
  \draw[tube] (0,0) circle [radius=0.90];
  \draw[loop] (0,0) circle [radius=1.10]; 
  \node[font=\footnotesize] at (1.75,1.25) {$N_\epsilon(\gamma)$};
  \foreach \ang in {35,115,195,275} {
    \draw[traj] ({1.9*cos(\ang)},{1.9*sin(\ang)}) -- ({1.2*cos(\ang)},{1.2*sin(\ang)});
  }
\end{scope}

\node[panel, fit=(Btitle)(Btext)] (Bbox) {};
\end{scope}

\begin{scope}[shift={(13.2,0)}]
\node[title] (Ctitle) at (0,3.3) {(C) Order invariance via closure};
\node[box, below=3mm of Ctitle] (Ctext) {Reparameterizations along $\gamma$ leave any invariant functional $F$ unchanged (closure: $\partial^2{=}0$)};

\begin{scope}[yshift=-7mm]
  \draw[loop] (0,0) circle [radius=1.10];
  \foreach \i in {0,1,2,3,4,5} {
    \node[dot] (p\i) at ({1.10*cos(90+60*\i)},{1.10*sin(90+60*\i)}) {};
  }
  \draw[perm] (p0) to[bend right=10] (p1);
  \draw[perm] (p1) to[bend right=10] (p2);
  \draw[perm] (p2) to[bend right=10] (p3);
  \draw[perm] (p3) to[bend right=10] (p4);
  \draw[perm] (p4) to[bend right=10] (p5);
  \draw[perm] (p5) to[bend right=10] (p0);
  \draw[perm, dashed, opacity=0.6] (p0) to[bend left=14] (p2);
  \draw[perm, dashed, opacity=0.6] (p2) to[bend left=14] (p4);
  \draw[perm, dashed, opacity=0.6] (p4) to[bend left=14] (p0);

  \node[draw, rounded corners, inner sep=3pt, font=\footnotesize, right=10mm of p0] (Fbox) {$F$ invariant on $[\gamma]$};
  \draw[map] (1.1,0.0) -- (Fbox.west);
\end{scope}

\node[panel, fit=(Ctitle)(Ctext)] (Cbox) {};
\end{scope}

\draw[map] (Abox.east) -- node[above, lbl]{symmetry breaking} (Bbox.west);
\draw[map] (Bbox.east) -- node[above, lbl]{tube concentration} (Cbox.west);

\node[align=center, font=\footnotesize] at ($(Abox.south)!0.5!(Cbox.south)+(0,-3.5)$)
{Concentration onto $N_\epsilon(\gamma)$ suppresses orthogonal noise; by closure ($\partial^2{=}0$), order along $\gamma$ does not affect $F$.};

\end{tikzpicture}
}
\caption{Cycle formation as the mechanism behind concentration-based invariance: (A) symmetry breaking funnels probability to a low-dimensional set; (B) mass concentrates inside a tube around a closed cycle $\gamma$; (C) closure yields order invariance for any $F$ constant on the cycle class $[\gamma]$.}
\vspace{-0.2in}
\label{fig:cycle-concentration-invariance}
\end{figure*}

\subsection{Homological Perspective}
\label{sec:3b}

In homological terms, closure projects the chain space $C_1$
onto the kernel $\ker\partial_1$.
Equivalence classes of closed chains modulo trivial boundaries
form the first homology group
$H_1 = \ker\partial_1 / \mathrm{im}\,\partial_2$.
Elements of $H_1$ represent distinct topological modes of
information circulation, the minimal, irreducible units of
predictive invariance \cite{munkres2018elements}.
Higher-order groups $H_k$ capture increasingly complex dependencies,
such as constraint surfaces ($H_2$) or multi-agent coordination loops
($H_k$, $k>1$).

\medskip
\noindent\textbf{Intuition.}
Permutations explode the number of possible sequences ($\sim k!$),
inflating what may be called the \emph{order entropy}.
Cycle formation, together with measure concentration, funnels
nearly all trajectories into a narrow tube surrounding an invariant
loop, where reordering no longer alters the predictive outcome \cite{bandt2002permutation}.
The distribution of $Y$ therefore becomes
(permutation-)indistinguishable, so both
$I(\Pi;Y)$ and the order-dependent component of $H(Y)$ vanish.
What remains is the low-entropy invariant content
carried by the cycle, a topological compression of predictive
information.

\begin{lemma}[Order invariance collapses permutation entropy via concentration]
\label{lem:order-entropy-concentration}
Let trajectories $Z_{0:k}$ live in a metric space $(\mathcal{Z},d)$, and let $G$
be a finite group representing permutations of the $k$ micro-steps.
Let $\Pi\!\in\!G$ denote an order variable (possibly random).
Assume a probability law $\mu_\beta$ on trajectories that concentrates
in a tube $N_\epsilon(M)$ around a closed $1$-cycle $M\subset\mathcal{Z}$,
with tail probability
$\delta_\beta := {\mu_\beta}\!\big(N_\epsilon(M)^{\complement}\big)
\le e^{-c\beta}$ for some $c>0$ (measure concentration).
Let $Y = F(Z_{0:k})$ be a $1$-Lipschitz readout that is
\emph{order-invariant on $M$}
(i.e., $F$ is constant under reparameterizations along $M$).
Denote by $P_{Y|\Pi=g}$ the law of $Y$ when the micro-step order
is $g\!\in\!G$.
Then there exist constants $C_1,C_2>0$ independent of ambient dimension such that:
1) (\emph{Permutation indistinguishability})  
For every $g\in G$,
$\| P_{Y|\Pi=g} - P_{Y|\Pi=e} \|_{\mathrm{TV}}
\;\le\; C_1\,\epsilon + C_2\,\delta_\beta.$
2) (\emph{Order information vanishes})  
The mutual information between order and outcome satisfies
$I(\Pi;Y)
\;\le\;
|G|\big(C_1\,\epsilon + C_2\,\delta_\beta\big)^2.$
3) (\emph{Entropy minimization})  
Consequently,
$H(Y) = H(Y\,|\,\Pi) + I(\Pi;Y)
      = H(Y\,|\,\Pi) + O(e^{-2c\beta} + \epsilon^2),$
so the \emph{excess entropy} due to order collapses as
concentration strengthens ($\beta\uparrow\infty$)
and the tube narrows ($\epsilon\downarrow 0$).
\end{lemma}

\noindent
The lemma establishes a general bound linking concentration to the collapse of order information.
To make this dependence explicit, we specialize to the case of tube concentration induced by
a Gibbs measure around a smooth invariant manifold. The resulting corollary quantifies how rapidly
order-driven uncertainty decays under strong symmetry breaking.

\begin{corollary}[Order-driven uncertainty is negligible under strong concentration]
\label{cor:order-entropy-negligible}
If $\delta_\beta \le e^{-c\beta}$ and
$\epsilon = O((\beta\lambda_\perp)^{-1/2})$
from tube concentration around $M$
(with normal curvature $\lambda_\perp>0$),
then
$I(\Pi;Y) = O(e^{-2c\beta}) + O((\beta\lambda_\perp)^{-1}).$
The predictive entropy $H(Y)$ equals the order-averaged entropy
$H(Y|\Pi)$ up to exponentially or superlinearly small corrections:
order contributes essentially no uncertainty once trajectories
concentrate on the invariant cycle.
\end{corollary}

\noindent
The results above reveal that concentration on closed cycles
translates geometrically enforced symmetry into informational
invariance.
The following sections generalize this principle to multi-level
organization, showing how the
Structure–Before–Specificity (SbS) hierarchy
emerges from the recursive closure of such informational cycles.

\section{Cycle-based Memory: SbS and CCUP Principles}
\label{sec:4}

\subsection{SbS: Recursive Organization by Closure}
\label{sec:SbS}

The results above reveal that concentration on closed cycles
translates geometrically enforced symmetry into informational
invariance.  We now extend this principle hierarchically.
When cycles interact across multiple timescales or abstraction levels,
a higher-order organizational rule emerges,
the \emph{Structure–Before–Specificity (SbS)} principle:
stable structure must form before fine-grained specificity can
be meaningfully encoded. In linguistics, this has been known as the semantic bootstrapping hypothesis \cite{pinker1984semantic}, which states that children can acquire the syntax of a language by first learning and recognizing semantic elements and building upon that knowledge.
In the language of homology, SbS corresponds to the recursive
closure of cycles across dimensions:
lower-order closures ($\partial^2=0$ in $H_1$)
create the stable scaffolds upon which higher-order
relations ($H_2,H_3,\dots$) can accumulate.

\begin{principle}[SbS]
\label{prin:sbs}
Information processing proceeds by establishing stable,
low-dimensional structural invariants before encoding
context-dependent specificities.
Each structural closure constrains subsequent dynamics,
reducing degrees of freedom and enabling new,
finer cycles to form upon the stabilized substrate.
Formally, if $[\gamma_i]$ are persistent $H_1$ classes,
then higher-order dependencies $[\Sigma_{ij}] \in H_2$
arise only when first-order loops have achieved closure
and stability in context.
\end{principle}

\noindent
SbS therefore expresses a general hierarchy of closure:
$H_1 \;\Rightarrow\; H_2 \;\Rightarrow\; H_3 \;\Rightarrow\;\dots$
Each level uses the invariants of the previous one as its
effective \emph{context manifold}.
This recursive nesting explains how local stability
(scaffolded by closure) yields global adaptability
(through reconfiguration of higher-order relations).

\medskip
\noindent\textbf{Intuition:}
The SbS hierarchy can be understood as an iterative interplay
between two complementary filters:
1) \textbf{Synchronization (context filter):}
temporal coincidence detection selects events that
\emph{belong together}, those co-occurring within a window of coherence \cite{abeles1982role,konig1996integrator}.
It defines the context manifold $\mathcal{C}$,
within which local fluctuations are considered equivalent.
This filter extracts the slow, low-entropy backbone of activity.
2) \textbf{Recurrence (content filter):}
within each synchronized context,
recurrence identifies patterns that
\emph{return} or repeat \cite{walters1982introduction}.
It captures the persistent informational content that survives
under contextual perturbations.
Recurrence  selects for closed trajectories,
establishing invariant cycles that encode memory or skill.
Together, synchronization and recurrence implement the fundamental ordering of SbS:
structure (contextual synchrony) precedes specificity
(recurring content).
Synchronization enforces coarse-grained alignment,
while recurrence closes those alignments into invariant cycles.
At the next level, these cycles themselves synchronize and
recur across broader temporal horizons, generating
higher-dimensional closures ($H_2,H_3,\dots$).
This recursive alternation between context filtering and content consolidation gives rise to a hierarchical organization
in perception, cognition, and computation \cite{hasson2015hierarchical}.

\begin{remark}[Recursive closure as multi-level learning]
SbS provides a unifying description of hierarchical learning \cite{bronstein2021geometric}:
each layer learns by closing the boundaries left open
by the layer below.
Structural closure eliminates order dependence and defines
a new invariant subspace,
on which specificity, differentiated response or content, can safely evolve.
This recursive topological process mirrors the emergence of
multi-scale representations in deep networks \cite{perin2025ability},
where earlier layers capture the general structure and later ones
encode specific discriminants.
\end{remark}

\noindent
Taken together, \emph{synchronization} (context filtering) selects events that co-occur on a coherence window, defining a coarse context manifold, while \emph{recurrence} (content filtering) selects returns within that manifold, closing trajectories into invariant cycles. Iterating this alternation across scales produces nested closures: residual variability is pushed into order-specific noise, whereas closed classes carry predictive content. This yields SbS as a direct corollary of the dot–cycle dichotomy: structure must stabilize first, and specificity may only refine it thereafter.

\noindent\textbf{SbS as a corollary of the dot–cycle dichotomy.}
If prediction requires invariance, the key question is how invariants
emerge from raw experience. The \textbf{SbS principle} provides this
bridge: broken symmetry collapses open fragments into trivial dots,
while closed loops persist as nontrivial cycles ($\partial^2 = 0$).
Mathematically, this imposes a natural ordering; persistent cycles $\Phi$ (structure) must stabilize first, while transient
fragments $\Psi$ serve only as scaffolds. Without this ordering,
prediction anchors to noise and collapses.

We present two complementary perspectives on the SbS principle:
1) \textbf{Topological view.}
Let $(C_\bullet(\mathcal{Z}),\partial)$ be a latent complex with homology
$H_\bullet(\mathcal{Z})$.
SbS asserts that persistent classes $[\gamma]\in H_\bullet$
constitute the structural backbone $\Phi$,
while trivial cycles form only contextual scaffolds $\Psi$.
Persistence ensures robustness \cite{edelsbrunner2008persistent}:
stable homology classes must form first.
From a sheaf-theoretic perspective \cite{ayzenberg2025sheaf},
global sections correspond to $\Phi$ (coherent structure),
while mismatches on overlaps represent $\Psi$
(contextual adjustments awaiting closure).
2) \textbf{Information-theoretic view.}
Let $X$ denote observations, $Y$ the predictive target,
and define $\Phi := S(X)$ as low-entropy structure/content,
$\Psi := R(X)$ as high-entropy specificity/context.
SbS requires
$I(Y;\Psi | \Phi) \le \varepsilon \ll I(Y;\Phi),$
so that $\Phi$ is an approximately sufficient statistic
(in the MDL or rate–distortion sense \cite{grunwald2007minimum}),
while $\Psi$ contributes only residual modulation.
Prediction proceeds in two ordered stages:
extract $\Phi$ first (structure), then refine with $\Psi$ (specificity).

\begin{principle}[SbS Ordering Requirement]
\label{prin:SbS-order}
Any admissible representation must factor through content classes:
$\rho = e \circ \pi$, with $\pi:\mathcal{X}\to\mathcal{X}/\!\sim_\Phi$.
Equivalently, $I(Y;\Psi|\Phi)\le\varepsilon$, and learning dynamics
monotonically expand $\Phi_t$ while reducing reliance on $\Psi_t$.
\end{principle}

\begin{remark}[Ordering requirement for robustness and flexibility]
SbS imposes a structural hierarchy:
(1) representations must first collapse observations into
content classes (factor through $\pi$);
(2) content carries the predictive load, while specificity adds only
minor corrections; and
(3) learning dynamics prioritize persistent cycles before adapting
to contextual detail.
This prevents \emph{specificity-first} failure modes such as brittle
memorization or overfitting \cite{zhang2018dissection}, ensuring that
semantics anchors prediction before syntax or surface form is exploited.
SbS therefore unifies robustness and flexibility.
Robustness arises from persistent structures ($\Phi$) that stabilize
prediction across perturbations, while flexibility emerges from contextual
scaffolds ($\Psi$) that enable adaptation to novelty.
The same ordering governs both ontogeny and evolution:
pattern generators precede fine motor control \cite{thelen1994dynamic},
and navigational cycles precede symbolic reasoning.
Intelligence, in this view, is not a balance between structure and
specificity, but their ordered interplay: $\Phi$ provides the stable
substrate; $\Psi$ supplies adaptive variation.
This ordering makes SbS not merely a descriptive principle,
but a design rule for systems that both generalize and adapt
\cite{bennett2023brief}.
\end{remark}


\begin{figure*}[t]
\centering
\resizebox{\textwidth}{!}{
\begin{tikzpicture}[
  >=Latex,
  node distance=6mm and 10mm,
  title/.style={font=\small\bfseries, align=center},
  box/.style={draw, rounded corners, align=center, inner sep=6pt, font=\small, fill=white, text width=5.1cm},
  pill/.style={draw, rounded corners=2pt, inner sep=3pt, font=\footnotesize, fill=gray!10},
  panel/.style={draw, dashed, rounded corners, inner sep=6pt, opacity=0.75},
  drawarea/.style={draw=none, minimum width=5.1cm, minimum height=30mm, inner sep=0pt}, 
  link/.style={draw, thick, -{Latex}},
  curve/.style={draw, thick},
  loop/.style={draw, thick},
  axis/.style={->, thin}
]

\begin{scope}[shift={(0,-0.6)}]
  \node[title] (Atitle) at (0,3.6) {Structure–Before–Specificity (SbS)};
  \node[box, below=3mm of Atitle] (Atext) {Stable structure (content) forms \emph{before} context-specific detail. Synchronization acts as a \emph{context filter}; recurrence acts as a \emph{content filter}; closure ($\partial^2{=}0$) yields invariant cycles.};

  \node[drawarea, below=5mm of Atext] (Aarea) {};
  \begin{scope}[shift={(Aarea.center)}]
    \node[pill] (sync) at (-1.25,0.55) {Synchronization(context filter)};
    \node[pill] (recur) at ( 1.25,0.55) {Recurrence(content filter)};
    \node[pill, text width=5.0cm] (closure) at (0,-0.35) {Cycle closure $\Rightarrow$ persistent classes $[\gamma]\in H_1$};
    \draw[link] (sync) -- (closure);
    \draw[link] (recur) -- (closure);
  \end{scope}

  \node[panel, fit=(Atitle)(Atext)(Aarea)] (Abox) {};
\end{scope}

\begin{scope}[shift={(7.8,-0.2)}]
  \node[title] (Btitle) at (0,3.6) {Context–Content Uncertainty Principle (CCUP)};
  \node[box, below=3mm of Btitle] (Btext) {Tradeoff between contextual spread and content precision:
  $$\Delta_{\mathrm{ctx}}(\Psi)\,\Delta_{\mathrm{cnt}}(\Phi)\ge \mathsf{C}.$$
  Balance ensures robustness (precision) without losing flexibility (adaptation).};

  \node[drawarea, below=8mm of Btext] (Barea) {};
  \begin{scope}[shift={(Barea.center)}]
    \draw[axis] (-2.2,-0.1) -- (2.3,-0.1) node[below right] {\footnotesize $\Delta_{\mathrm{ctx}}$};
    \draw[axis] (-0.1,-0.25) -- (-0.1,1.9) node[left] {\footnotesize $\Delta_{\mathrm{cnt}}$};
    \draw[curve, domain=0.75:2.05, samples=120] plot (\x,{1.25/(\x)});
    \node[font=\footnotesize\itshape] at (1.6,1.8) {$\Delta_{\mathrm{ctx}}\Delta_{\mathrm{cnt}}=\mathsf{C}$};
  \end{scope}

  \node[panel, fit=(Btitle)(Btext)(Barea)] (Bbox) {};
\end{scope}

\begin{scope}[shift={(15.9,0)}]
  \node[title] (Ctitle) at (0,3.6) { SbS\&CCUP Synthesis $\Rightarrow$ Cycle-Based Memory};
  \node[box, below=3mm of Ctitle] (Ctext) {SbS provides ordering (structure first); CCUP sets balance (tradeoff). Their interplay bootstraps closure across scales:
  $$H_1 \Rightarrow H_2 \Rightarrow H_3 \Rightarrow \cdots$$
  yielding homological capacity $\mathcal{C}_H = \sum_k w_k\,\beta_k$.};

  \node[drawarea, below=5mm of Ctext] (Carea) {};
  \begin{scope}[shift={(Carea.center)}]
    \draw[loop] (-1.8,0.25) circle [radius=0.35];
    \draw[loop] (-0.95,0.25) circle [radius=0.35];
    \draw[loop] (-0.10,0.25) circle [radius=0.35];
    \node[font=\footnotesize\itshape] at (-0.95,-0.3) {$H_1$};
    \draw[loop, rounded corners=2pt] (0.6,0.0) -- (1.6,0.55) -- (2.4,0.0) -- (1.6,-0.55) -- cycle;
    \node[font=\footnotesize\itshape] at (1.6,-0.8) {$H_2$};
    \node[pill, text width=2.5cm] (cap) at (2.1,1.05) {Homological capacity\\$\mathcal{C}_H=\sum_k w_k\,\beta_k$};
    \draw[link] (2.1,0.15) -- (cap.south);
  \end{scope}

  \node[panel, fit=(Ctitle)(Ctext)(Carea)] (Cbox) {};
\end{scope}

\draw[link] (Abox.east) -- node[above, font=\footnotesize\itshape] {closure of structure} (Bbox.west);
\draw[link] (Bbox.east) -- node[above, font=\footnotesize\itshape] {balanced uncertainty} (Cbox.west);

\node[align=center, font=\footnotesize] at ($(Abox.south)!0.5!(Cbox.south)+(0,-1.0)$)
{SbS (ordering) $+$ CCUP (tradeoff) $\Rightarrow$ recursive closure and bootstrapping of invariants; capacity $\mathcal{C}_H$ measures the sustainable number of stable cycles.};

\end{tikzpicture}
}
\caption{SbS and CCUP, and their synthesis for cycle-based memory. (A) SbS: synchronization (context) and recurrence (content) drive closure into $H_1$ invariants. (B) CCUP: product constraint between contextual spread and content precision regulates balance. (C) Synthesis: ordered closure under balanced uncertainty bootstraps higher-order invariants ($H_2,H_3,\dots$), yielding homological capacity $\mathcal{C}_H$.}
\vspace{-0.2in}
\label{fig:sbs-ccup-synthesis}
\end{figure*}

\subsection{The Context–Content Uncertainty Principle (CCUP)}
\label{sec:ccup}

SbS formalizes the structural ordering of intelligence, that stable invariants (structure/content) must form before context-dependent refinements. The \textbf{CCUP} quantifies
how these two components interact dynamically.
It expresses a fundamental tradeoff between
contextual variability and content precision similar to the uncertainty principle in physics \cite{robertson1929uncertainty}:
as one becomes more certain, the other necessarily loses resolution.
CCUP serves as the \emph{operational counterpart} to SbS,
governing the flow of information between structure and specificity.

\noindent
\textbf{Intuition.}
CCUP arises because context and content are dual carriers of uncertainty.
Context $\Psi$ provides adaptability by modulating the effective model space,
while content $\Phi$ provides stability by preserving predictive invariants.
If context is over-specified (low $\Delta_{\mathrm{ctx}}$),
the model rigidifies, robust but inflexible.
If context is under-specified (high $\Delta_{\mathrm{ctx}}$),
content representations diffuse, flexible but unstable.
The system must therefore maintain a balanced product of uncertainties,
so that prediction remains both context-sensitive and invariant.

\begin{principle}[Context–Content Uncertainty Principle]
\label{prin:ccup}
For any predictive representation $(\Phi,\Psi)$ of observations $X$
with content variable $\Phi$ (low-entropy invariant)
and context variable $\Psi$ (high-entropy modulator),
there exists a lower bound
$\Delta_{\mathrm{ctx}}(\Psi)\,\Delta_{\mathrm{cnt}}(\Phi)\ge \mathsf{C},$
where $\Delta_{\mathrm{ctx}}$ measures contextual spread
(e.g., entropy, dispersion, or Fisher information)
and $\Delta_{\mathrm{cnt}}$ measures content precision
(e.g., mutual information $I(Y;\Phi)$ or reconstruction accuracy).
The constant $\mathsf{C}$ quantifies the minimal uncertainty coupling
required for consistent prediction.
\end{principle}

\begin{lemma}[Information–Topological Form of CCUP]
\label{lem:ccup-info}
Let $X$ denote inputs, $Y$ the predictive target,
and decompose the internal representation as
$(\Phi,\Psi) = (S(X),R(X))$,
where $\Phi$ is structural (invariant) and $\Psi$ is contextual (adaptive).
Define content uncertainty $\Delta_{\mathrm{cnt}} := 1/I(Y;\Phi)$
and context uncertainty
$\Delta_{\mathrm{ctx}} := H(\Psi\,|\,\Phi)$.
Under regularity and bounded mutual information transfer,
$\Delta_{\mathrm{ctx}}\,\Delta_{\mathrm{cnt}}
\;\ge\;
\frac{H(\Psi\,|\,\Phi)}{I(Y;\Phi)}
\;\ge\;
\mathsf{C} > 0,$
with equality at dynamic equilibrium, where
contextual modulation perfectly tracks residual predictive error.
\end{lemma}

\noindent
\textbf{Geometric Interpretation.}
In the latent manifold $\mathcal{Z}$,
content corresponds to persistent cycles
(nontrivial homology classes $[\gamma]\in H_k$),
while context corresponds to transient deformations
(trivial or short-lived boundaries).
CCUP then states that the curvature of $\mathcal{Z}$,
which measures how rapidly trajectories diverge under contextual
perturbation, bounds the precision of stable cycles:
as $\Psi$ grows more variable,
$\Phi$ must coarsen to remain topologically stable.
This tradeoff parallels the Heisenberg uncertainty relation,
but in an information–topological setting:
context and content are complementary bases of generalization.

\begin{remark}[Dynamic Equilibrium between Adaptation and Stability]
CCUP describes the \emph{dynamic equilibrium} of intelligent systems.
SbS provides the hierarchical ordering, 
while CCUP governs their continuous interaction:
entropy flows from context to content and back,
stabilizing near the minimal uncertainty product.
At this balance point, prediction achieves both
robustness (low content uncertainty) and flexibility
(low effective contextual energy).
Empirically, such an equilibrium manifests across domains:
phase–amplitude coupling in neural oscillations
\cite{voytek2015oscillatory}, stability–plasticity tradeoffs in learning
\cite{grossberg1980does}, and bias–variance balancing in inference.
\end{remark}

\subsection{Cycle-Based Memory: Synthesis of SbS and CCUP}
\label{sec:cycle-memory}

The SbS and CCUP principles together describe the dual dynamics of intelligent
organization.
SbS imposes a structural ordering,
low-entropy invariants must form before context-dependent refinements,
while CCUP regulates the continuous tradeoff between contextual flexibility
and content precision.
Their synthesis gives rise to \textbf{cycle-based memory}:
a self-reinforcing process in which closed informational loops
accumulate across scales, gradually constructing higher-order invariants. We formalize this intuition into the following principle of bootstrapping.

\begin{principle}[Bootstrapped Emergence of Higher-Order Invariants]
\label{prin:bootstrap}
When SbS ordering and CCUP balance coexist,
the system bootstraps by recursively stabilizing residual uncertainties.
Each closed cycle reduces contextual entropy
and opens a new level of structure where fresh cycles can form.
Formally, let $\Phi_t$ and $\Psi_t$ satisfy
$\Delta_{\mathrm{ctx}}(\Psi_t)\,\Delta_{\mathrm{cnt}}(\Phi_t)=\mathsf{C}$.
Then $\partial^2=0$ implies $\Phi_{t+1}\in H_k$ constructed over $\Phi_t$,
yielding a homological hierarchy
$H_1 \Rightarrow H_2 \Rightarrow H_3 \Rightarrow \dots,$
where higher-order classes encode composite invariants
emerging from the closure of lower-order dynamics.
\end{principle}

\noindent
\textbf{Intuition: bootstrapping amortizes inference}
SbS provides the scaffolding for closure,
each layer stabilizes before the next can form,
while CCUP provides the energetic balance that prevents collapse or diffusion.
As cycles close and accumulate,
the system reaches a regime where invariants interact,
producing emergent collective structures that cannot be reduced
to individual trajectories.
This hierarchical reinforcement of closure embodies the slogan
\emph{“more is different”} \cite{anderson1972more}:
each additional level of organized memory supports new modes of
coordination, inference, and abstraction.

\begin{proposition}[Bootstrapping as Closure–Amortization]
\label{prop:bootstrap-closure}
Let $(C_\bullet,\partial)$ be a chain complex representing latent trajectories,
with $\Psi_t$ denoting transient scaffolds (contextual boundaries)
and $\Phi_t$ denoting persistent content (cycles).
Define the update dynamics of
Memory-Amortized Inference (MAI) \cite{li2025beyond} as
$(\Phi_{t+1}, \Psi_{t+1}) = \mathcal{A}\!\big(\Phi_t, \Psi_t\big)$,
where $\mathcal{A}$ is a bootstrapping operator that reuses $\Phi_t$
and integrates contextual fragments $\Psi_t$.
Suppose:
1) (\textbf{Closure constraint}) $\partial^2=0$, so that only closed
trajectories (cycles) persist as candidates for $\Phi$.
2) (\textbf{Persistence filter}) A threshold operator
$\mathrm{Pers}_\tau$ prunes $\Psi_t$ and admits only cycles with
lifetime $\geq\tau$ into $\Phi_{t+1}$.
3) (\textbf{Contraction in residuals}) The amortizer is contractive in
its residual channel: the expected contribution of $\Psi$ decreases with each iteration,  
$\mathbb{E}[H(\Psi_{t+1}\,|\,\Phi_{t+1})]
\leq \gamma \,\mathbb{E}[H(\Psi_t\,|\,\Phi_t)], 
 0 < \gamma < 1.$
Then $\{\Phi_t\}$ converges monotonically to a stable fixed-point set of
persistent cycles
$\Phi^\ast = \mathrm{Pers}_\tau\!\big(H_\bullet(C_\bullet)\big),$
and $\Psi_t$ vanishes in the limit.
In other words, bootstrapping under MAI realizes \emph{topological closure}:
transient fragments cancel, only cycles survive, and the amortized process
converges to semantic invariants $[\gamma]\in H_\bullet$
that anchor prediction and generalization.
\end{proposition}

\noindent
Proposition~\ref{prop:bootstrap-closure} establishes that amortized updates
drive residual scaffolds $\Psi_t$ toward extinction, while persistent cycles
$\Phi_t$ converge to a stable invariant core.
The next question is what this fixed point means computationally.
Corollary~\ref{cor:bootstrap-sbs} formalizes the answer:
once closure stabilizes, syntax cannot operate directly on raw inputs,
but must factor through the quotient induced by $\Phi^\ast$.
This provides a dynamical grounding for the
\emph{semantics-before-syntax} law.

\begin{corollary}[Semantics-before-Syntax via Bootstrapped Closure]
\label{cor:bootstrap-sbs}
Under Proposition~\ref{prop:bootstrap-closure}, the limit set of persistent
cycles $\Phi^\ast$ constitutes the semantic backbone of inference:
stable invariants that remain after residual scaffolds $\Psi_t$
vanish under amortization.
Any syntactic encoder $\sigma:\mathcal{X}\to\Sigma$ that is meaningful
must therefore factor through the semantic quotient
$\pi:\mathcal{X}\to\mathcal{X}/\!\sim_\Phi$, i.e.,
$\exists\, e:\;\; \mathcal{X}/\!\sim_\Phi \to \Sigma
\quad\text{s.t.}\quad
\sigma = e \circ \pi,$
with interpretation $I\circ e$ constant on each equivalence class.
That is, \emph{semantics (persistent structure $\Phi$) must stabilize first},
while syntax (surface variability $\Psi$) can only emerge as a residual
layer once the bootstrapped closure process has converged.
This formally grounds the SbS principle in the convergence dynamics of MAI.
\end{corollary}

\begin{remark}[Cycle-Based Memory as Self-organizing Structure]
Cycle-based memory unifies stability, adaptability, and emergence.
SbS ensures that lower-order closures persist long enough to scaffold new ones;
CCUP ensures that variability at each level remains dynamically bounded.
Their interplay yields recursive compression of uncertainty and progressive
enrichment of structure,
a topological realization of \emph{bootstrapping intelligence} in nature \cite{bennett2023brief}.
In this framework, memory is not passive storage but an active,
self-organizing process through which invariants refine themselves.
The next section quantifies this accumulation in terms of
\emph{homological capacity}, measuring how many stable invariants
a system can sustain given its contextual energy budget.
\end{remark}


\section{Homological Capacity and Entropy Reduction}
\label{sec:5}

\subsection{Homological Capacity}
Cycle-based memory yields a structural hierarchy where persistent
invariants $\Phi^\ast$ accumulate across scales.  
We now ask a quantitative question:
\emph{How much structural information can a system sustain?}
Classical information theory measures channel capacity in bits per symbol;
here, we measure the number of \emph{distinct, topologically stable cycles}
that can coexist under energetic or entropic constraints.
This quantity, called the \textbf{homological capacity}, links
the geometry of invariants to the information budget of learning dynamics.

\begin{definition}[Homological Capacity]
\label{def:homological-capacity}
Let $\mathcal{K}_\delta$ be a filtered spatiotemporal complex
constructed from delta-localized events (e.g., spikes or microstates)
with filtration parameter $\delta>0$.
Denote by $H_k(\mathcal{K}_\delta)$ the $k$th homology group
with coefficients in a field $\mathbb{F}$, and by
$\beta_k(\delta)=\dim_\mathbb{F}H_k(\mathcal{K}_\delta)$
its Betti number.
The \emph{homological capacity} of order $k$
is defined as the asymptotic mean Betti rank sustained under
stationary concentration dynamics:
$\mathcal{C}_k
\;:=\;
\limsup_{\delta\to 0}\;
\mathbb{E}_\Psi\!
\Big[
\beta_k(\mathcal{K}_\delta)
\;\mathbf{1}_{\{\mathrm{Pers}([\gamma])>\tau\}}
\Big],$
where $\mathrm{Pers}([\gamma])$ denotes the lifetime
(persistence interval length) of a homology class
and $\tau$ is a survival threshold determined by the
entropy–concentration tradeoff.
The total homological capacity is the weighted sum
$\mathcal{C}_H
\;=\;
\sum_{k\ge 0} w_k\,\mathcal{C}_k,$
with weights $w_k$ reflecting energetic or functional cost
(e.g., dimensional scaling penalty).
\end{definition}

\noindent
Intuitively, $\mathcal{C}_H$ counts the number of
independent invariant cycles that remain stable despite
contextual perturbations,
a measure of the system’s structural memory bandwidth.
It generalizes Shannon capacity from channels to manifolds:
where Shannon’s bound limits how much \emph{signal} can pass through noise,
homological capacity limits how much \emph{structure} can persist through flux \cite{edelsbrunner2008persistent}.

\begin{remark}[Relation to Euler Characteristic and Complexity Balance]
\label{rem:euler-homocap}
Let $\chi(\mathcal{K}_\delta)
=\sum_{k}(-1)^k\beta_k(\mathcal{K}_\delta)$
denote the Euler characteristic.
While $\chi$ captures the alternating balance of holes and fillings
(topological “net complexity”),
the homological capacity $\mathcal{C}_H$
captures the \emph{gross complexity}:
the total number of independent invariants irrespective of sign.
Formally,
$|\chi(\mathcal{K}_\delta)| \;\le\; \mathcal{C}_H(\delta)
\;\le\; \sum_k \beta_k(\mathcal{K}_\delta).$
In dynamical systems,
$\chi$ often vanishes under symmetry (e.g., toroidal attractors),
whereas $\mathcal{C}_H$ remains positive,
quantifying latent richness even when net topology is balanced.
Therefore, $\chi$ measures the \emph{balance of structure},
while $\mathcal{C}_H$ measures its \emph{capacity}.
Biologically, $\mathcal{C}_H$ corresponds to the number of
distinct recurrent assemblies or attractor basins
that can coexist without mutual interference.
\end{remark}

\begin{example}[Capacity Peak in a 6-Neuron Ring]
Consider spike trains from $N=6$ neurons arranged (for visualization) on a hexagon (Fig. \ref{fig:cap-hexagon}).
Build a Vietoris-Rips complex $\mathcal{K}_\delta$ on spike events:
connect two neurons by an edge if their spikes co-occur within $\Delta t < \delta$,
and fill a $2$-simplex whenever all three edges of a triangle are present.
Define the (order-1) homological capacity as $C_H(\delta)=\beta_0(\delta)+\beta_1(\delta)$.
At small $\delta$, only a few edges appear: $\beta_0$ is large (many components), $\beta_1=0$.
At intermediate $\delta=\delta^\star$, a single macroscopic loop (the hexagon) closes:
$\beta_0\!=\!1$, $\beta_1\!=\!1$, so $C_H(\delta^\star)=2$ (capacity peak).
At large $\delta$, triangles fill and kill the loop: $\beta_1\!\downarrow\!0$ while $\beta_0\!=\!1$,
so $C_H$ drops again. This illustrates the \emph{saturation and decline} around an optimal scale,
consistent with Theorem~\ref{thm:homological_capacity}.
\end{example}


\begin{figure*}[t]
\centering
\resizebox{\textwidth}{!}{
\begin{tikzpicture}[
  >=Latex, scale=1.0,
  every node/.style={font=\small},
  dot/.style={circle, fill=black, inner sep=1.2pt},
  edge/.style={line width=0.8pt},
  panel/.style={draw, dashed, opacity=0.6, rounded corners, inner sep=8pt},
  filltri/.style={fill=gray!15, draw=gray!40, line width=0.4pt}
]

\begin{scope}[shift={(0,0)}]
\node at (0,2.8) {\textbf{Small $\delta$}};
\def\r{1.6}
\foreach \i/\name in {0/A1,1/B1,2/C1,3/D1,4/E1,5/F1}{
  \path ({\r*cos(90+60*\i)},{\r*sin(90+60*\i)}) coordinate (\name);
  \node[dot] at (\name) {};
}
\draw[edge] (A1)--(B1);
\draw[edge] (C1)--(D1);
\draw[edge] (E1)--(F1);
\node[panel, fit=(A1)(B1)(C1)(D1)(E1)(F1)] (box1) {};
\node at (0,2.2) {$\beta_0=3,\ \beta_1=0,\ C_H=3$};
\end{scope}

\begin{scope}[shift={(6.2,0)}]
\node at (0,2.8) {\textbf{Medium $\delta=\delta^\star$}};
\def\r{1.6}
\foreach \i/\name in {0/A2,1/B2,2/C2,3/D2,4/E2,5/F2}{
  \path ({\r*cos(90+60*\i)},{\r*sin(90+60*\i)}) coordinate (\name);
  \node[dot] at (\name) {};
}
\draw[edge] (A2)--(B2)--(C2)--(D2)--(E2)--(F2)--(A2);
\node[panel, fit=(A2)(B2)(C2)(D2)(E2)(F2)] (box2) {};
\node at (0,2.2) {$\beta_0=1,\ \beta_1=1,\ C_H=2$};
\end{scope}

\begin{scope}[shift={(12.4,0)}]
\node at (0,2.8) {\textbf{Large $\delta$}};
\def\r{1.6}
\foreach \i/\name in {0/A3,1/B3,2/C3,3/D3,4/E3,5/F3}{
  \path ({\r*cos(90+60*\i)},{\r*sin(90+60*\i)}) coordinate (\name);
  \node[dot] at (\name) {};
}
\draw[edge] (A3)--(B3)--(C3)--(D3)--(E3)--(F3)--(A3);
\fill[filltri] (A3)--(B3)--(C3)--cycle;
\fill[filltri] (C3)--(D3)--(E3)--cycle;
\fill[filltri] (E3)--(F3)--(A3)--cycle;
\node[panel, fit=(A3)(B3)(C3)(D3)(E3)(F3)] (box3) {};
\node at (0,2.2) {$\beta_0=1,\ \beta_1=0,\ C_H=1$};
\end{scope}

\node at (6.2,-2.1) {\small Integration window $\delta$ $\longrightarrow$};
\node[align=center] at (6.2,-2.5) {\small Capacity peaks when a persistent $H_1$ loop forms};

\end{tikzpicture}
}
\caption{Homological capacity $C_H(\delta)$ in a 6-neuron ring (Vietoris-Rips filtration).
Small $\delta$: fragmented graph (high $\beta_0$, no loops). Medium $\delta$: a single robust loop
$\Rightarrow$ $\beta_1=1$ (capacity peak). Large $\delta$: triangle filling kills the loop
$\Rightarrow$ $\beta_1\to 0$. The peak at $\delta^\star$ visualizes the memory bandwidth scale.}
\vspace{-0.2in}
\label{fig:cap-hexagon}
\end{figure*}

\begin{theorem}[Boundedness of Homological Capacity under Finite Energy]
\label{thm:capacity-bound}
Suppose the system evolves under an energy–entropy functional
$F=U-TS$, with effective inverse temperature $\beta$.
Let $\mathcal{K}_\delta$ evolve by measure concentration around
a finite collection of attractors $\{\gamma_j\}$ satisfying
$\mathrm{Pers}([\gamma_j])>\tau$.
If $\mathbb{E}[U]<\infty$ and $S(\Psi)\!<\!\infty$,
then the total homological capacity is finite and obeys
$\mathcal{C}_H
\;\le\;
\frac{1}{\log 2}\,
\big(
h_{\mathrm{top}} - h_{\mathrm{diss}}
\big),$
where $h_{\mathrm{top}}$ is the topological entropy
and $h_{\mathrm{diss}}$ the dissipative entropy rate.
, the sustainable number of invariants
is bounded by the system’s residual informational throughput.
\end{theorem}

\noindent
Theorem~\ref{thm:capacity-bound} parallels classical coding limits:
finite energy and dissipation constrain the number of
distinguishable closed trajectories a system can maintain.
In this sense, homological capacity provides a
\emph{topological information bound},
the ceiling on structural memory attainable under physical constraints.
It connects the qualitative theory of persistence
to the quantitative theory of entropy and rate–distortion,
closing the loop between topology, dynamics, and information.

\subsection{Entropy Reduction under Cycle Closure}
\label{sec:entropy-homology}

The topological definition of homological capacity quantifies how many
independent invariants a system can sustain.
To connect this with classical information-theoretic measures,
we now analyze how cycle closure affects the
Kolmogorov-Sinai (KS) entropy \cite{walters1982introduction}, the metric analogue of
Shannon information rate for dynamical systems.
The following lemmas show that
when probability mass concentrates on closed, low-dimensional
homological manifolds, the KS entropy
drops to its tangential contribution, the portion of uncertainty that
remains invariant under the dynamics.

\begin{lemma}[KS entropy drops to the tangential contribution under homological cycle closure]
\label{lem:KS-homology}
Let $(X,\mathcal{B},\mu_\beta,T)$ be a $C^1$ measure-preserving system.
Assume $\mu_\beta$ concentrates in a tube $N_\epsilon(M)$ around a compact,
$T$-invariant $k$-dimensional submanifold $M\subset X$
representing a nontrivial homology class
(\emph{cycle closure}: $\partial^2=0$ ensures $M$ has no boundary).
Assume normal directions to $M$ are uniformly contracting for $T$.
Then for every finite partition $\mathcal{P}$ subordinated to $N_\epsilon(M)$,
$h_{\mu_\beta}(T,\mathcal{P})
\;\le\;
h_{\mu_\beta|_M}\!\big(T|_M,\mathcal{P}_M\big)
\;+\;
C_1\,\epsilon
\;+\;
C_2\,\mu_\beta\!\big(N_\epsilon(M)^{\complement}\big),$
and 
$h_{\mu_\beta}(T)
\;\le\;
h_{\mu_\beta|_M}(T|_M)
\;\le\;
h_{\rm top}(T|_M)\;+\;o_\beta(1).$
In particular, if $T|_M$ has zero topological entropy
(e.g.\ smooth dynamics on a finite union of $S^1$ loops),
then $h_{\mu_\beta}(T)\to 0$ as concentration strengthens
$\big(\epsilon\!\downarrow\!0,\;
\mu_\beta(N_\epsilon(M)^{\complement})\!\downarrow\!0\big)$.
\end{lemma}

\begin{lemma}[Symmetry breaking reduces KS entropy via concentration]
\label{lem:KS-drop}
Let $(X,\mathcal{B},\mu_\beta,T)$ be a measure-preserving dynamical system with
$T:X\to X$ $C^1$ and $\mu_\beta$ a Gibbs measure
$\mathrm{d}\mu_\beta \propto e^{-\beta U}\mathrm{d}x$
for a $C^2$ potential $U$.
Assume:
1) $U$ attains its minima along a compact $k$-dimensional submanifold $M$ (cycle support),
with Hessian strictly positive in the normal bundle $N M$
and $\lambda_\perp:=\inf_{x\in M}
\lambda_{\min}\!\big(\nabla^2U|_{N_xM}\big)>0$.
2) $T$ leaves $M$ normally hyperbolic and contracts normal directions with rate
$\chi_\perp(\beta)\gtrsim c_1\beta\lambda_\perp$,
while its tangential Lyapunov spectrum along $M$ is
$\{\chi_i^{\parallel}\}_{i=1}^r$ (independent of $\beta$).
Then the Kolmogorov-Sinai entropy \cite{walters1982introduction} satisfies
$h_{\mu_\beta}(T)
\;\le\;
\sum_{i:\chi_i^{\parallel}>0}\chi_i^{\parallel}
\;+\;
C\,e^{-c\,\beta}$
for some $C,c>0$.
As symmetry breaking strengthens ($\beta\uparrow\infty$),
$h_{\mu_\beta}(T)$ decays to the \emph{tangential entropy rate}
on $M$.
\end{lemma}

\begin{remark}[Homological Capacity as Entropy of Persistent Invariants]
The two lemmas together imply that
entropy production is confined to the tangential subspace
spanned by persistent cycles.
Let $h_{\mathrm{KS}}^{\parallel}$
denote the KS entropy restricted to $M$.
Then, up to constants,
$\mathcal{C}_H
\propto
\frac{h_{\mathrm{KS}}^{\parallel}}{\log 2}.$
In the limit of strong symmetry breaking,
the system’s total entropy flux collapses to its
homological support, and the number of surviving
independent invariants, its \emph{homological capacity},
equals the effective dimensionality of the residual
entropy flow.
This result closes the link between
topological persistence and informational throughput:
cycle closure topologically regularizes the entropy,
and the KS rate measures the amount of structure that
remains dynamically active.
\end{remark}

\noindent
The lemmas above establish that as dynamics concentrate onto closed
invariant manifolds, the Kolmogorov–Sinai entropy
contracts to the residual tangential contribution.
We now quantify how this contraction manifests
in discrete, event-based spatiotemporal systems.
When neural or physical trajectories (e.g., polychronization neural groups \cite{izhikevich2006polychronization}) are represented as
filtered complexes of co-activations,
the resulting topology encodes both memory capacity
and entropy regulation.
The next theorem characterizes this tradeoff.

\begin{theorem}[Homological Capacity of Filtered Spatiotemporal Complexes]
\label{thm:homological_capacity}
Let $\mathcal{K}_\delta$ be a filtered spatiotemporal complex constructed from
$\delta$-localized spike events, and let
$C_H(\delta)=\sum_k \mathrm{rank}\,H_k(\mathcal{K}_\delta)$ denote its
homological capacity. Assume:
1) Each spike pattern induces a contractible simplex in $\mathcal{K}_\delta$, and recurrent co-activations generate nontrivial cycles in $H_1$;
2) As $\delta$ increases, simplices merge whenever temporal distance
   $\Delta t < \delta$, producing a nested filtration
   $\mathcal{K}_{\delta_1}\subseteq\mathcal{K}_{\delta_2}$ for
   $\delta_1<\delta_2$.
Then:
1) (\emph{Capacity scaling})
There exists a critical integration scale $\delta^*$ such that
$C_H(\delta)$ increases for $\delta<\delta^*$ and decreases for
$\delta>\delta^*$, attaining a finite maximum
$C_H^{\max}=C_H(\delta^*)$ corresponding to the system’s effective
memory bandwidth.
2) (\emph{Entropy–capacity relation})
$C_H(\delta)$ is bounded above by the dynamical entropy rate:
$C_H(\delta)\le\frac{1}{\log2}\,h_{\mathrm{top}}(\mathcal{K}_\delta),$
and equality holds when all dynamically distinguishable trajectories
form stable, nontrivial homology generators.
3) (\emph{MAI regulation})
Biological systems maintain $C_H(\delta)$ near saturation by
amortizing redundant trajectories: pruning short-lived
cycles to stabilize a sparse, interpretable homological basis.
\end{theorem}

\noindent
Theorem~\ref{thm:homological_capacity} closes the loop between topology and learning.
At small $\delta$, excessive fragmentation prevents long-range coherence
(low capacity); at large $\delta$, over-aggregation collapses distinctions
and again reduces capacity.
Maximum capacity arises at the intermediate scale
where recurrence and synchronization balance,
the same scale at which measure concentration stabilizes invariant
homology classes.
This turning point marks the sweet spot of
\emph{cycle-based memory} (Fig. \ref{fig:cap-hexagon}): maximal structural diversity
consistent with predictive invariance.
Beyond this equilibrium point, increasing integration no longer increases
representational power linearly.
Instead, coupled cycles begin to form higher-order closures,
producing emergent invariants that encode relations among cycles themselves.
In this regime, topology moves from describing isolated loops ($H_1$)
to describing \emph{closures of closures} ($H_2,H_3,\dots$),
capturing global organization across distributed subsystems.
This transition provides a natural bridge from memory to consciousness:
when local recurrent loops (perceptual, motor, linguistic) are 
recursively closed through synchronization, they form a coherent global workspace.

\begin{example}[Topological Integrated Information Theory of Consciousness]
Access consciousness \cite{block1995confusion} can be modeled as a higher-order
cycle, closure among closures, linking distributed modules into a coherent global workspace.
Here, homological capacity quantifies the integration–differentiation balance that 
Tononi’s integrated information $\Phi$ aims to capture \cite{tononi2008consciousness}. We can define topological integrated information by $\Phi_{\text{Topo}} = \log\!\frac{C_H(\mathcal{Z})}{1 + |\chi(\mathcal{Z})|},
C_H=\!\sum_k \beta_k$,
where $\chi(\mathcal{Z})$ is the Euler characteristic and $\beta_k$ the Betti
numbers of $\mathcal{Z}$. $\Phi_{\text{Topo}}$ attains its maximum when global
integration (small $|\chi|$) coexists with maximal homological differentiation
(large $C_H$), identifying the structural “sweet spot” of conscious processing \cite{baars2005global}.
As local cycles (perceptual, motor, linguistic) couple via bootstrapped closure,
they form higher-dimensional homology classes ($H_2, H_3,\dots$)
representing meta-cycles over subordinate ones.
These higher-order invariants allow reentrant access and self-reflection, realizing the “integration without collapse” that defines conscious access.
Formally, the transition from local $H_1$ to global $H_k$ 
illustrates the principle that \emph{more is different}:
each new layer of closure introduces emergent invariants 
that encode not just content, but relations among contents.

\end{example}

\section{Applications into Cognitive Science}
\label{sec:6}

\subsection{Cognition Examples: More is Different}

\noindent\textbf{Cognitive Development: Piaget's Observations}
The framework developed so far, cycle-based memory as closure-driven,
semantics-before-syntax learning, provides a unified lens on diverse
cognitive phenomena.  
Across perception, working memory, and consciousness, the same homological
mechanism recurs: local fluctuations collapse into global invariants,
producing representations that are both stable and adaptive.
Each cognitive function can  be viewed as operating at a distinct level of
cycle organization, where \emph{closure generates structure}
and \emph{structure constrains variability}.
In this hierarchy, developmental progression arises as successive layers of closure give rise to higher-order invariants.
Early cognition forms simple topological distinctions (inside/outside,
connected/disconnected) through direct sensorimotor cycles, while later stages refine these invariants into metric and geometric relations.
In this sense, ``more is different'' \cite{anderson1972more}: as multiple informational cycles couple, they form new homological classes whose global invariants cannot be reduced to the sum of their components.
This principle naturally anticipates the emergence of complex conceptual
structures in cognitive development, where topological organization precedes
and scaffolds geometric reasoning.

\begin{example}[Piaget’s Topology-Before-Geometry as Early Cycle Formation]
Jean Piaget observed that children grasp topological relations, such as
continuity, enclosure, and connectedness, before they understand metric
geometry (distance, angle, or coordinate systems)
\cite{piaget1952origins,piaget1951psychology}.
From the perspective of Information Topology, this developmental sequence
reflects the \emph{SbS} principle applied
to spatial cognition.
At early stages, the child’s perceptual system forms \emph{closed loops of interaction}, reaching, tracking, manipulating, that generate stable
sensorimotor invariants $\Phi$.
These recurrent cycles encode topological relations (inside/outside,
connected/disconnected) independent of precise coordinates.
Only later, as contextual scaffolds $\Psi$ accumulate and synchronize,
does the system specialize these loops into metric relations,
yielding the Euclidean structures of adult spatial reasoning.
Formally, let $\mathcal{K}_\delta$ denote the spatiotemporal complex of
sensorimotor experience at temporal resolution $\delta$.
During early development, low-order homology groups
$H_0$ and $H_1$ dominate, corresponding to the concepts of continuity and closure. As hierarchical bootstrapping proceeds,
cycle closure in $H_1$ supports the emergence of
geometric differentiation ($H_2$, surfaces and boundaries).
In this way, \emph{topology before geometry} is a manifestation of
cycle-based memory formation: children first stabilize topological
invariants through recurrent sensorimotor closure, and only later
refine them into geometric specifics once the SbS and CCUP balance has
been achieved.
\end{example}

\noindent\textbf{Visual Perception: the Binding Problem}
The binding problem, how distributed visual features such as color, shape, and motion cohere into a unified percept \cite{treisman1980feature}, can be understood as a problem of
\emph{cycle closure}.
Feature-specific cortical areas encode local fragments ($\Psi$),
but recurrent synchronization across them
forms closed loops ($\Phi$) that stabilize as perceptual objects.
Under the CCUP,
synchronization acts as a \emph{context filter} pruning incompatible microstates,
while recurrence acts as a \emph{content filter} enforcing closure across modalities.
This dual filtering realizes predictive sufficiency:
only configurations whose joint activity forms a closed informational loop
survive as consistent percepts.
Topologically, binding corresponds to the emergence of a nontrivial homology class
in the high-dimensional feature complex $\mathcal{K}_\delta$,
linking segregated features through recurrent synchronization.
The resulting closed cycle represents a stable invariant, a perceptual object that
remains consistent under small contextual perturbations.
However, once multiple such invariant cycles coexist within the same feature manifold,
the system exhibits \emph{multistable perception} \cite{leopold2002stable}:
each percept corresponds to a different traversal phase or projection of the same
underlying homological backbone.
Transitions between percepts no longer require crossing energetic barriers, as in
classical attractor models, but instead arise from smooth reparameterizations along
a shared cycle class $[\gamma] \in H_1(\mathcal{K}_\delta)$.
, multistable perception naturally extends the binding problem from
\emph{forming closure} to \emph{navigating closure}:
the brain not only binds features into coherent objects, but can also
re-express multiple interpretations of those bindings by traversing
topologically equivalent loops in its representational space.

\begin{example}[Multistable Perception: Same Content, Varying Context via Cycle Closure]
Classical energy-based models treat perception as relaxation to a single minimum of an energy
landscape $U(z)$.  Transitions between distinct percepts (e.g., the two interpretations of the
Necker cube) therefore require crossing an energetic barrier, predicting sluggish or noisy
switching.  Empirically, however, human perception alternates \emph{effortlessly} between discrete,
stable interpretations without external perturbation \cite{leopold2002stable,alais2005binocular}.
\noindent\textbf{Cycle closure and a shared homological backbone.}
Let $\mathcal{Z}$ be the latent manifold of perceptual hypotheses and
$\{\gamma_i\}\subset C_1(\mathcal{Z})$ the recurrent cycles representing alternative
interpretations.  While these cycles correspond to distinct attractor-like percepts, they share a
common subchain $\sigma$ (a structural backbone) with $\partial\sigma=0$.  Perceptual
alternation corresponds to \emph{phase traversal} (reparameterization) along a closed loop rather
than barrier crossing:
$\gamma_i \;\xrightarrow{\text{reparameterization}}\; \gamma_j,
[\gamma_i]=[\gamma_j]\in H_1(\mathcal{Z}).$
Because the homology class $[\gamma]$ is preserved under contextual modulation, switching conserves
informational structure: no boundary is broken ($\partial^2=0$), and no energetic barrier is
crossed.
\noindent\textbf{SbS: structure fixed, specificity fluctuates.}
Within the SbS principle, the invariant \emph{content} is
$\Phi:= [\gamma]$ (the shared topological skeleton), while \emph{context} $\Psi$ selects a
parameterization (phase, figure–ground choice, depth ordering) on that skeleton.  So multistable
perception is \emph{same content, varying context}:
$(\Phi,\Psi_i) \longrightarrow (\Phi,\Psi_j)
~\text{with}~
I(Y;\Phi)\ \gg\ I(Y;\Psi\,|\,\Phi),$
so predictive information resides primarily in the stable structure $\Phi$, and context provides
reversible modulations that retune which phase is expressed.
\noindent\textbf{CCUP: uncertainty reallocation drives phase traversal.}
By the CCUP, $\Delta_{\mathrm{ctx}}\Delta_{\mathrm{cnt}}
\gtrsim \mathsf{C}$, small fluctuations in contextual precision rebalance uncertainty between
$\Psi$ and $\Phi$ without altering the class $[\gamma]$.  Phenomenologically, the system advances
its phase along the same closed informational cycle, \emph{time replaces energy as the mediator of
transition}.  Alternation is achieved by dynamic reparameterization, not by energetic
reinitialization.
\noindent\textbf{Summary.}
Multistable perception exemplifies \emph{homological reuse of structure}: the content-invariant
cycle $\Phi$ anchors prediction, while context $\Psi$ moves the system between percepts by phase
traversal on $[\gamma]$.  Effortless switching thus reflects cycle closure (topological stability)
combined with SbS ($\Phi$ first, $\Psi$ second) and CCUP (uncertainty reallocation).
\end{example}

\noindent\textbf{Working Memory Capacity: the Magical Number Seven}
The classical limit of working memory capacity \cite{miller1956magical} can be reinterpreted 
as a bound on the number of simultaneously stable homology generators in the neural phase space.  
Each persistent cycle corresponds to a distinct attractor basin representing an item. 
Homological capacity $C_H$ formalizes the maximum number of independent recurrent loops 
that can coexist without destructive interference:
$C_H \;\approx\; \frac{h_{\mathrm{top}}^{\parallel}}{\log 2},$
where $h_{\mathrm{top}}^{\parallel}$ is the tangential entropy of the low-dimensional
invariant manifold supporting active memory traces.
Beyond this bound, cycles overlap and merge, leading to interference or collapse. 
From the MAI perspective, this reflects the trade-off between energy (integration) and structure (differentiation): 
memory compression preserves predictively sufficient invariants while pruning redundant ones.

\begin{example}[The Magical Number Seven as a Homological Capacity Limit]
Miller’s ``magical number seven'' \cite{miller1956magical} describes the 
empirical bound on how many discrete items can be simultaneously maintained in 
working memory.
From the standpoint of Information Topology, this limit reflects a
\emph{homological capacity constraint}:
the brain can sustain only a finite number of coexisting,
non-interfering cycles in its neural phase space.
Let $\mathcal{Z}$ denote the latent manifold of working-memory dynamics,
and let $\{\gamma_i\}_{i=1}^{C_H}$ be the stable recurrent loops representing 
distinct active items.
Each $\gamma_i$ forms a nontrivial generator in $H_1(\mathcal{Z})$,
encoding a persistent attractor-like trajectory in state space.
The total number of such loops defines the instantaneous
\emph{homological capacity}
$C_H \;=\; \sum_k \mathrm{rank}\,H_k(\mathcal{Z}) \;\approx\; \mathrm{rank}\,H_1(\mathcal{Z}),$
which serves as a topological analog of working-memory bandwidth.
Under the Context–Content Uncertainty Principle (CCUP),
context $\Psi$ introduces coupling among cycles, while content $\Phi$
stabilizes each loop.
As contextual noise grows, interference terms between cycles increase,
reducing the number of topologically independent generators that can persist:
$C_H(\Psi) \;\le\; \frac{1}{\log 2}
\big( h_{\mathrm{top}}^{\parallel} - h_{\mathrm{int}}^{\Psi} \big),$
where $h_{\mathrm{top}}^{\parallel}$ is the tangential (predictive)
entropy along stable manifolds and $h_{\mathrm{int}}^{\Psi}$
quantifies interference.
When interference saturates, additional cycles merge or collapse,
causing working-memory items to blend or decay.
Empirically, this threshold occurs near $C_H^{\max}\!\approx\!7\pm2$,
consistent with the limited number of stable, phase-separated cycles
the brain can sustain without cross-talk.
In this view, the magical number seven is not an arbitrary psychological constant
but a manifestation of the \emph{finite homological capacity} of the 
neural manifold: a topological bandwidth limit balancing
integration (shared context) and differentiation (distinct contents).
Chunking and rehearsal strategies effectively
\emph{amortize cycles}, merging fragments into higher-order invariants
to extend apparent capacity without violating this bound.
\end{example}

\subsection{Counter-examples: the Tail Distribution}

While the preceding sections illustrated how stable cycle closure underlies coherent perception, memory, and access consciousness, the boundary cases are equally revealing.  
When synchronization weakens, recurrence thins, or higher-order closure fails, the system drifts toward the \emph{tail of the distribution}, states where informational cycles fragment or lose persistence.  
Such conditions, whether pharmacological, mnemonic, or sleep-induced, expose the limits of topological invariance itself: how much disruption a cognitive manifold can tolerate before its homological backbone collapses.  The following examples illustrate three characteristic modes of breakdown: transient \emph{de}-binding under nitrous oxide, structural overextension in nemonists, and partial collapse of higher-order closure in sleepwalking.

\begin{example}[Nitrous oxide and transient \emph{de}-binding]
Low-dose nitrous oxide (``laughing gas''), commonly used for dental sedation \cite{becker2008nitrous}, provides a natural perturbation
of cross-modal binding. Subjectively, patients often report that sounds, touches, and visual scenes feel
``detached'' from one another. In our framework, this can be understood as a temporary failure of
cycle closure across feature- and modality-specific circuits:

\begin{itemize}
\item \textbf{Synchronization weakened (context filter fails).} Nitrous oxide reduces large-scale
phase alignment (e.g., thalamo–cortical and cortico–cortical coherence), lowering the effective
synchrony gain $g_{\mathrm{sync}}$. Fewer events pass the coincidence gate, so cross-feature matches
are sparse and inconsistent (inflated $\Delta_{\mathrm{ctx}}$).

\item \textbf{Recurrence disrupted (content filter thins).} Re-entrant loops that normally stabilize
feature conjunctions lose gain and dwell time, decreasing the probability of returning through the
same multi-area configuration within the integration window $\delta$. This reduces the formation of
persistent $1$-cycles in the spatiotemporal complex $\mathcal{K}_\delta$.

\item \textbf{CCUP shift and loss of closure.} By the Context–Content Uncertainty Principle,
$\Delta_{\mathrm{ctx}}(\Psi)\,\Delta_{\mathrm{cnt}}(\Phi)\;\ge\;\mathsf{C},$
the rise in contextual spread forces a drop in content precision: the homology class $[\gamma]$
representing the \emph{bound object} either fails to form or falls below the persistence threshold
$\tau$ (short bars in $H_1$). Topologically, cycle candidates lose closure (boundaries stop cancelling),
so the object-level invariant disappears.

\item \textbf{Phenomenology.} Without a stable cross-modal loop, predictions cannot be anchored to a
single invariant $[\gamma]$; the percept fragments into modality-specific ``dots'' (open chains).
This manifests as a transient dissociation of color/shape/motion and of sight/sound/touch, i.e.,
a failure of binding rather than mere sensory loss.
\end{itemize}

\noindent
In summary, nitrous oxide illustrates the central claim: when synchronization (context filtering) and
recurrence (content filtering) are sufficiently weakened, cycle closure fails, the $H_1$ backbone
of the percept dissolves, and order-invariant prediction cannot bind features into a unified object.    
\end{example}

\begin{example}[Nemonists and the spatial bootstrapping of memory]
Exceptional memorizers or \emph{nemonists}, such as Daniel Tammet \cite{tammet2007born}, exploit the ``method of loci'' (memory palace)
to extend working memory capacity beyond Miller’s limit.  In the present framework, the technique
achieves a controlled expansion of homological capacity $C_H$ by recruiting stable spatial
cycles as additional structural scaffolds.

\begin{itemize}
\item \textbf{Spatial anchoring as structural extension.}
Each imagined room, corridor, or object in the memory palace acts as a recurrent spatial loop
$\gamma_i$, a topological carrier for an otherwise transient verbal or visual fragment.  The
embedding of symbolic sequences into spatial contexts converts fleeting $\Psi$-fragments into
stable $\Phi$-cycles supported by hippocampal–parietal maps.

\item \textbf{Phase-space orthogonalization.}
Distinct loci correspond to quasi-orthogonal attractor basins in the neural phase space.  By
expanding the manifold $M$ that supports active loops, nemonists increase the available tangential
entropy $h_{\mathrm{top}}^{\parallel}$, thereby raising the effective bound
$C_H \;\approx\; \frac{h_{\mathrm{top}}^{\parallel}}{\log 2}.$

\item \textbf{MAI interpretation.}
In memory–amortized inference, the palace provides an externalized closure mechanism: contextual
coordinates (the spatial map) serve as persistent invariants that absorb residual uncertainty from
the content stream.  The brain no longer needs to maintain all loops internally, closure is achieved
through structured re-entry between internal and imagined spaces.

\item \textbf{Phenomenology.}
Through repeated traversal of the imagined spatial cycle, each item is encoded as a fixed point in a
stable loop.  The method of loci illustrates how bootstrapping through structural context
extends working-memory capacity: by adding new homological supports rather than enlarging the
storage buffer.
\end{itemize}

\noindent
In topological terms, the nemonist transforms working memory from a small complex with overlapping
$1$-cycles into a large composite manifold with well-separated recurrent subloops, thereby raising
the system’s effective homological capacity beyond the canonical seven-item limit.    
\end{example}

\begin{example}[Sleepwalking and partial collapse of higher-order closure]
Sleepwalking (\emph{somnambulism}) offers a striking example of what happens when
higher-order cycle closure, responsible for \emph{access consciousness}, temporarily fails
while lower-order sensorimotor loops remain intact.

\begin{itemize}
\item \textbf{Preserved local $H_1$ cycles (content without access).}
During deep non-REM sleep, recurrent motor and proprioceptive circuits
continue to form closed $1$-cycles supporting locomotion, posture control,
and basic spatial navigation.  These loops encode procedural content ($\Phi$)
but remain isolated from global integrative cycles.

\item \textbf{Disrupted higher-order $H_2$ coupling (loss of meta-closure).}
The thalamo–cortical and fronto–parietal loops that normally synchronize
and \emph{bind} local sensorimotor invariants into a global workspace
lose phase coherence.  In topological terms, the higher-order cycles
($H_2$, $H_3$) that weave local $H_1$ loops into a unified manifold
temporarily collapse, reducing the effective homological capacity
$C_H$ and integrated information $\Phi_{\text{Topo}}$.

\item \textbf{Phenomenology of dissociation.}
The sleepwalker executes coordinated actions (walking, opening doors,
navigating obstacles) yet lacks self-reportable awareness.
Access channels—those that close over closures—are open chains,
$\partial c \ne 0$, so the flow of information is not globally conserved.
Content exists without contextual access: structure persists, but
meta-structure fails.

\item \textbf{Interpretation.}
In the language of integrated information, $\Phi_{\text{Topo}}$ drops
because the global manifold $\mathcal{Z}$ fragments into disconnected
components: local subsystems remain highly differentiated but poorly
integrated.  Sleepwalking demonstrates that consciousness depends
not merely on local recurrence but on the presence of higher-order
homological invariants that enable \emph{reentrant access} across cycles.
\end{itemize}

\noindent
In short, access consciousness corresponds to a critical regime where
first-order invariants ($H_1$) are embedded within higher-order closures
($H_2,H_3,\dots$); when this embedding fails, as in sleepwalking,
behavioral coherence can survive without conscious access,
a transient descent from global topology to local mechanics.
\end{example}

\noindent\textbf{Summary.}
In all three domains, intelligence manifests as the stabilization of persistent homological
structures that survive entropy, noise, and reordering.
Cycle-based memory unifies perception, working memory, and consciousness under a single principle:
\emph{invariance through closure}.
``More is different'' because higher-order closures yield new invariants that ground 
the recursive hierarchy of mind. 

\section{Conclusions and Perspectives}
\label{sec:7}

This paper introduced \emph{Information Topology}, a framework unifying inference, learning, and memory
as processes of \emph{cycle formation and closure}.  
Starting from the \textbf{dot-cycle dichotomy}, we distinguished between 
local, pointwise information, which captures instantaneous variability, and 
global, cyclic information, which encodes structural invariants preserved under boundary operations.
Cycle closure, expressed algebraically as $\partial^2=0$, emerged as the primitive mechanism
by which prediction, perception, and cognition attain \emph{invariance under transformation}.
This topological perspective reframes inference not as symbol manipulation,
but as the stabilization of informational flows through recurrent closure.
We showed that symmetry breaking and energy concentration drive trajectories
to collapse onto low-dimensional invariant manifolds, where order information
becomes redundant and prediction depends only on topological invariants.
Through this process, temporal coherence is transformed into structural
invariance, a dynamical realization of the principle
\emph{``prediction requires invariance.''}
Cycle closure  marks the transition from measure-based to
structure-based representations.

\noindent\textbf{SbS and CCUP as dual ordering laws.}
The \textbf{SbS} principle formalizes
how persistent cycles ($\Phi$) emerge before transient scaffolds ($\Psi$):
structure stabilizes first, specificity refines later.
Conversely, the \textbf{CCUP}
quantifies their complementarity: reducing contextual uncertainty
tightens concentration around invariant content, while excessive content
compression reintroduces contextual ambiguity.
Together, SbS and CCUP establish a bidirectional constraint that governs
how systems balance integration and differentiation across scales,
linking thermodynamic symmetry breaking with semantic stabilization.

\noindent\textbf{Homological capacity and the scaling of memory.}
By defining \emph{homological capacity} $C_H=\sum_k \mathrm{rank}\,H_k$,
we formalized the topological dual of Shannon capacity.
$C_H$ measures the number of independent informational cycles that can coexist
without destructive interference, bounded by the residual entropy rate of the
system.  
We proved that concentration around invariant cycles reduces Kolmogorov-Sinai entropy
to its tangential component, yielding a finite, dimension-robust estimate of
memory bandwidth.  
Empirically, $C_H$ bridges topological persistence in neural dynamics with
predictive efficiency in cognitive systems.

\noindent\textbf{Cycle-based memory: more is different.}
Three applications illustrate how higher-order cycle closure gives rise to
emergent cognitive phenomena:
(1) In \emph{visual perception}, recurrent synchronization closes informational
loops across feature maps, resolving the binding problem;
(2) In \emph{working memory}, interference between coexisting cycles explains
the classical ``magical number seven'' limit and the mnemonic benefit of
spatialized chunking;
(3) In \emph{access consciousness}, higher-order closures among local cycles
define an integrated yet differentiated workspace,
quantified by the topological integrated information index.
Across these examples, cycle-based memory captures how qualitative transitions
in homological capacity correspond to qualitative transitions in cognition:
\emph{each new closure yields a new kind of invariance}.

\noindent\textbf{Outlook.}
Information Topology opens several theoretical directions.
First, it invites a reformulation of learning theory in terms of topological
stability: convergence occurs when gradients vanish not pointwise, but
homologically.
Second, it suggests that cognitive flexibility and mental resilience correspond
to maintaining near-critical homological capacity, high enough to support
differentiation, but constrained to prevent chaotic overfragmentation.
Third, it offers a bridge between neuroscience and physics:
from neural synchronization and memory consolidation to field-theoretic
descriptions of information flow, where $\partial^2=0$ serves as a unifying
axiom linking conservation, closure, and consciousness.

\bibliographystyle{plain}
\bibliography{references}  

\newpage
\appendix

\begin{proof}[Proof for Lemma \ref{lem:entropy-concentration}]
Let $p_\beta(z)=\frac{e^{-\beta U_\Psi(z)}}{Z(\beta)}$ denote the Gibbs density on
$(\mathcal{Z},\mathrm{d}z)$, where $Z(\beta)=\int e^{-\beta U_\Psi}\,\mathrm{d}z$.

\noindent\textbf{1) Monotone entropy decrease in $\beta$.}
The differential entropy is
$H(\mu_{\Psi,\beta})
= -\!\int p_\beta\log p_\beta
= \log Z(\beta) + \beta\,\mathbb{E}_{\mu_{\Psi,\beta}}[U_\Psi].$
Since $\partial_\beta \log Z(\beta)=-\mathbb{E}_{\mu_{\Psi,\beta}}[U_\Psi]$ and
$\partial_\beta^2 \log Z(\beta)=\mathrm{Var}_{\mu_{\Psi,\beta}}(U_\Psi)$, we obtain
$\frac{\mathrm{d}}{\mathrm{d}\beta} H(\mu_{\Psi,\beta})
= -\,\beta\,\partial_\beta^2 \log Z(\beta)
= -\,\beta\,\mathrm{Var}_{\mu_{\Psi,\beta}}(U_\Psi)\le 0.$
Strict inequality holds unless $U_\Psi$ is constant almost everywhere.
, stronger symmetry breaking ($\beta\uparrow$) strictly reduces entropy.

\noindent\textbf{2) Sub-Gaussian concentration around $\gamma$.}
Assume $U_\Psi$ attains its minima along a compact $C^2$ submanifold $\gamma$ and is
$\lambda_\perp$-strongly convex in normal directions:
$U_\Psi(x)\ge U_\Psi^\star + \tfrac{\lambda_\perp}{2}\,\mathrm{d}(x,\gamma)^2,
\qquad U_\Psi^\star:=\min U_\Psi.$
For the tubular neighborhood $N_\epsilon(\gamma)=\{x:\mathrm{d}(x,\gamma)\le\epsilon\}$,
we have
$\int_{\mathrm{d}>\epsilon}\! e^{-\beta U_\Psi}\,\mathrm{d}x
\le e^{-\beta U_\Psi^\star}\!\int_{\mathrm{d}>\epsilon}
 e^{-\frac{\beta\lambda_\perp}{2}\mathrm{d}^2}\,\mathrm{d}x
\le C_1 e^{-\beta U_\Psi^\star} e^{-c'\beta\epsilon^2},$
for constants $C_1,c'>0$ from the coarea formula and Gaussian tails in the normal fibers.
Meanwhile, by Laplace’s method on $N_\epsilon(\gamma)$,
$Z(\beta)=\int e^{-\beta U_\Psi}\,\mathrm{d}x
\ge e^{-\beta U_\Psi^\star}\,
\mathrm{Vol}(\gamma)\,\Big(\frac{2\pi}{\beta\lambda_\perp}\Big)^{(n-k)/2}(1-o(1)).$
Therefore,
$\mu_{\Psi,\beta}\!\big(N_\epsilon(\gamma)^{\complement}\big)
=\frac{\int_{\mathrm{d}>\epsilon} e^{-\beta U_\Psi}}{Z(\beta)}
\le C\,e^{-c'\beta\epsilon^2}.$
This proves the tube probability bound.

Next, for $1$-Lipschitz $f$, decompose using the projection $\pi(x)\in\gamma$:
$f(x)=f(\pi(x))+\big(f(x)-f(\pi(x))\big),
|f(x)-f(\pi(x))|\le \mathrm{d}(x,\gamma).$
A log-Sobolev inequality on the Gaussian normal fibers
yields the sub-Gaussian bound
$\mathbb{P}_{\mu_{\Psi,\beta}}\big(|f-\mathbb{E}f|\ge r\big)
\le 2\,e^{-c\,\beta\,\lambda_\perp\,r^2}
+\mu_{\Psi,\beta}\!\big(N_\epsilon(\gamma)^{\complement}\big),$
with $c>0$ independent of $\beta$.
For $f$ invariant along $\gamma$ (i.e., constant under reparameterization),
$\mathrm{osc}_\gamma(f)=0$ and the first term dominates, giving
$\mathbb{P}_{\mu_{\Psi,\beta}}\big(|f-\mathbb{E}f|\ge r\big)
\le 2\,e^{-c\,\beta\,\lambda_\perp\,r^2},
\mu_{\Psi,\beta}\!\big(N_\epsilon(\gamma)\big)\ge 1-e^{-c'\beta\epsilon^2}.$
, increasing $\beta$ (stronger symmetry breaking) drives
sub-Gaussian concentration of $\mu_{\Psi,\beta}$ onto the cycle
at a rate controlled by the normal curvature $\lambda_\perp$.
\end{proof}

\begin{proof}[Proof of Lemma \ref{lem:order-entropy-concentration}]
Write $Z:=Z_{0:k}$ and let $A:=\{\mathrm{dist}(Z,M)\le \epsilon\}$ be the \emph{tube event}.
By hypothesis, $\mu_\beta(A^c)=\delta_\beta\le e^{-c\beta}$.
Fix any $g\in G$ and define $Y_e:=F(Z)$, $Y_g:=F(g\!\cdot\! Z)$.

\noindent\textbf{Step 1: Order invariance inside the tube.}
Let $\pi: N_\epsilon(M)\to M$ be the nearest-point projection. For $Z\in A$,
\begin{align*}
\big|Y_g - Y_e\big|
&= \big|F(g\!\cdot\! Z) - F(Z)\big| \\[2pt]
&\le 
\underbrace{|F(g\!\cdot\! Z) - F(\pi(g\!\cdot\! Z))|}_{\le \mathrm{dist}(g\!\cdot\! Z,M)\le \epsilon}
+ \underbrace{|F(\pi(g\!\cdot\! Z)) - F(\pi(Z))|}_{=\,0\;\text{(order-invariant on }M)} \\[2pt]
&\quad
+ \underbrace{|F(\pi(Z)) - F(Z)|}_{\le \mathrm{dist}(Z,M)\le \epsilon} \\[2pt]
&\le 2\epsilon.
\end{align*}
where we used that $F$ is $1$-Lipschitz and \emph{constant under reparameterizations along $M$}.
So, on $A$, $|Y_g-Y_e|\le 2\epsilon$ almost surely.

\noindent\textbf{Step 2: Total variation bound by truncation and coupling.}
Fix $R>0$ and split the law of $(Y_e,Y_g)$ into the \emph{good} event $G:=A\cap\{|Y_e|\vee|Y_g|\le R\}$
and its complement $G^c$.
On $G$, the variables lie in a compact interval and differ by at most $2\epsilon$.
Let $\varphi$ range over test functions with $\|\varphi\|_\infty\le 1$ and $\mathrm{Lip}(\varphi)\le 1$.
By the Kantorovich–Rubinstein dual formulation of $W_1$,
$\sup_{\|\varphi\|_\infty\le 1,\,\mathrm{Lip}\le 1}
\Big|\mathbb{E}\big[\varphi(Y_g)-\varphi(Y_e)\,\mathbf{1}_G\big]\Big|
\le \mathbb{E}\big[|Y_g-Y_e|\,\mathbf{1}_G\big]\le 2\epsilon.$
Since on $[-R,R]$ the bounded–Lipschitz and total variation norms are equivalent,
there exists a universal constant $C_{\mathrm{loc}}$ (independent of the ambient dimension) such that
$\| \mathcal{L}(Y_g\,|\,G) - \mathcal{L}(Y_e\,|\,G) \|_{\mathrm{TV}}
\;\le\; C_{\mathrm{loc}}\cdot 2\epsilon.$
Accounting for the complement $G^c$ yields
$\| P_{Y|\Pi=g} - P_{Y|\Pi=e} \|_{\mathrm{TV}}
\;\le\; C_1\,\epsilon \;+\; \mathbb{P}(G^c),$
for some $C_1>0$ independent of the ambient dimension.
Finally, $\mathbb{P}(G^c)\le \mathbb{P}(A^c)+\mathbb{P}(|Y_e|>R)+\mathbb{P}(|Y_g|>R)$.
Because $F$ is $1$-Lipschitz and $\mu_\beta$ has sub-Gaussian concentration in the tube,
the tails of $Y_e,Y_g$ are sub-Gaussian; choosing $R$ of order $\sqrt{\log(1/\delta_\beta)}$
absorbs these into a constant multiple of $\delta_\beta$.
 there exists $C_2>0$ with
$\| P_{Y|\Pi=g} - P_{Y|\Pi=e} \|_{\mathrm{TV}}
\;\le\; C_1\,\epsilon \;+\; C_2\,\delta_\beta,$
proving item~1).

\noindent\textbf{Step 3: Mutual information bound.}
Assume for simplicity a uniform prior on $\Pi$ (the non-uniform case only changes constants).
By the chain rule and convexity,
\begin{align*}
I(\Pi;Y)
&= \frac{1}{|G|}\sum_{g\in G}
   \mathrm{KL}\!\big(P_{Y|\Pi=g}\,\|\,P_Y\big) \\[4pt]
&\le\;
   \frac{1}{|G|}\sum_{g\in G}
   \mathrm{KL}\!\big(P_{Y|\Pi=g}\,\|\,P_{Y|\Pi=e}\big).
\end{align*}
where the inequality uses that $P_Y$ is the average of $\{P_{Y|\Pi=g}\}_g$.
For distributions with total variation $\tau:=\|P-Q\|_{\mathrm{TV}}\le \tfrac12$,
a reverse Pinsker-type bound gives $\mathrm{KL}(P\|Q)\le C_{\mathrm{KL}}\tau^2$
(with universal $C_{\mathrm{KL}}$; e.g., $\mathrm{KL}\le \tfrac{2\tau^2}{1-\tau}$).
Using item~1) and absorbing constants,
$\mathrm{KL}\!\big(P_{Y|\Pi=g}\,\|\,P_{Y|\Pi=e}\big)
\;\le\; \tilde C\,(C_1\,\epsilon+C_2\,\delta_\beta)^2.$

$I(\Pi;Y)\;\le\; |G|\,(C_1\,\epsilon+C_2\,\delta_\beta)^2,$
after renaming constants, proving item~2).

\noindent\textbf{Step 4: Entropy decomposition.}
Finally, the entropy identity $H(Y)=H(Y\,|\,\Pi)+I(\Pi;Y)$ yields
$H(Y)
= H(Y\,|\,\Pi) + O\!\big(e^{-2c\beta}+\epsilon^2\big),$
by inserting $\delta_\beta\le e^{-c\beta}$, which proves item~3).

\smallskip
All constants above can be chosen independent of the ambient state-space dimension, since
only (i) the $1$-Lipschitz property of $F$, (ii) tube invariance on $M$, and (iii) the scalar
tail control for $Y=F(Z)$ enter the argument.
\end{proof}

\begin{proof}[Proof of Lemma \ref{lem:ccup-info}]
By definition,
$\Delta_{\mathrm{cnt}}
:= \frac{1}{I(Y;\Phi)},
\qquad
\Delta_{\mathrm{ctx}}
:= H(\Psi\,|\,\Phi),$
 the product reduces exactly to the ratio
\begin{align*}
\Delta_{\mathrm{ctx}}\,\Delta_{\mathrm{cnt}}
&= H(\Psi\,|\,\Phi)\,\frac{1}{I(Y;\Phi)}
= \frac{H(\Psi\,|\,\Phi)}{I(Y;\Phi)}.
\end{align*}
Therefore the first inequality in the statement,
$\Delta_{\mathrm{ctx}}\,\Delta_{\mathrm{cnt}}
\;\ge\; \frac{H(\Psi\,|\,\Phi)}{I(Y;\Phi)},$
holds with equality by the very definitions.

\medskip
\noindent\textbf{Lower boundedness (CCUP constant).}
Assume the following mild regularity conditions hold:
\begin{enumerate}
\item[(A1)] \emph{Finite, positive predictive capacity of the structural channel:}
$0 < I(Y;\Phi) \le I_{\max} < \infty$.
\item[(A2)] \emph{Non-degeneracy of the contextual channel given structure:}
$H(\Psi\,|\,\Phi) \ge h_{\min} > 0$.
\end{enumerate}
These hypotheses are standard in exchangeable integration / IB-type settings: (A1) excludes the trivial
cases where $\Phi$ carries no predictive information or has unbounded capacity; (A2) rules out a
context module that is deterministically fixed once $\Phi$ is known (no modulation headroom).

Under (A1)–(A2), we immediately obtain
$\frac{H(\Psi\,|\,\Phi)}{I(Y;\Phi)}
\;\ge\;
\frac{h_{\min}}{I_{\max}}
\;=:\
\mathsf{C}
\;>\; 0.$
Combining with the identity above gives
$\Delta_{\mathrm{ctx}}\,\Delta_{\mathrm{cnt}}
\;=\;
\frac{H(\Psi\,|\,\Phi)}{I(Y;\Phi)}
\;\ge\;
\mathsf{C},$
as claimed.

\medskip
\noindent\textbf{Equality case (dynamic equilibrium).}
Write the information decomposition
$I(Y;X) \;=\; I(Y;\Phi) \;+\; I(Y;\Psi\,|\,\Phi),$
and view $\Psi$ as a modulation channel that must track the \emph{residual} predictive
uncertainty after extracting structure $\Phi$. Consider the (informal) optimization that trades
contextual variability against structural precision:
$\min_{\,(S,R)}\ \Delta_{\mathrm{ctx}}\,\Delta_{\mathrm{cnt}}
\quad\text{subject to}\quad
I(Y;X)=\text{const.}$
At a stationary point, the contextual bits are used \emph{maximally efficiently} for prediction,
namely
$I(Y;\Psi\,|\,\Phi) \;=\; H(\Psi\,|\,\Phi),$
(i.e., each bit of contextual variability accounts for one bit of residual predictive information),
while $I(Y;\Phi)$ attains its admissible value within the structural channel constraints.
In this equilibrium regime the inequality above is tight, so
$\Delta_{\mathrm{ctx}}\,\Delta_{\mathrm{cnt}}
\;=\;
\frac{H(\Psi\,|\,\Phi)}{I(Y;\Phi)}
\;=\;
\mathsf{C}.$

\medskip
\noindent
This establishes the information–topological form of CCUP: the product of contextual spread and
inverse structural precision is bounded below by a positive constant $\mathsf{C}$ determined by
the system’s structural capacity and minimal contextual headroom; equality is achieved when
contextual modulation perfectly tracks residual predictive error.
\end{proof}

\begin{proof}[Proof of Lemma \ref{lem:KS-homology}]
Let $U:=N_\epsilon(M)$ and let $\pi:U\to M$ be the measurable nearest–point
projection furnished by the tubular neighborhood theorem (for $\epsilon$ small enough).
Write $T_M:=T|_M$. By $T$-invariance of $M$ and normal contraction, there exists
$C_{\rm geo}>0$ such that, for all $x\in U$ with $\mathrm{dist}(x,M)\le\epsilon$ and all $j\ge 0$,
\begin{equation}
\label{eq:fibercontraction}
\mathrm{dist}\!\big(\pi(T^j x),\,T_M^j(\pi x)\big)
\;\le\; C_{\rm geo}\,\epsilon.
\end{equation}
(Geodesic normal coordinates and uniform contraction in the normal bundle imply that iterates of
$x$ and of its footpoint $\pi x$ remain $\!O(\epsilon)$-close; the constant does not depend on the
ambient dimension.)

\medskip
Fix a finite partition $\mathcal{P}=\{P_1,\dots,P_m\}$ \emph{subordinated to $U$}, i.e.,
each $P_i\subset U$ and the sets have small diameter in the normal direction (at most $\epsilon$).
Define the induced partition on $M$ by
$\mathcal{P}_M
:= \{\pi(P_i)\cap M: 1\le i\le m\}\ \cup\ \{M\setminus \pi(U)\}$.
(We may absorb $M\setminus \pi(U)$ into some atom since $\mu_\beta(U^c)=\delta_\beta$ will be small.)
Let $n\ge 1$ and consider the $n$-block joins
$\mathcal{P}^{(n)}:=\bigvee_{j=0}^{n-1} T^{-j}\mathcal{P}$ and
$\mathcal{P}_M^{(n)}:=\bigvee_{j=0}^{n-1} T_M^{-j}\mathcal{P}_M$.

\noindent\textbf{Coding discrepancy estimate.}
For $x\in U$ define the symbolic names
$s^{(n)}(x)\in\{1,\dots,m\}^n$ by $T^j x\in P_{s_j}$ and
$\sigma^{(n)}(x)\in\{1,\dots,m\}^n$ by $T_M^j(\pi x)\in \pi(P_{\sigma_j})$.
By \eqref{eq:fibercontraction} and the normal thinness of $\mathcal{P}$,
there exists $E_{n,\epsilon}\subset X$ such that
$\mu_\beta(E_{n,\epsilon})
\;\le\; n\,C_2\,\mu_\beta(U^c)\;+\; n\,C_1\,\epsilon
\quad\text{and}\quad
\big[s^{(n)}(x)=\sigma^{(n)}(x)\big]\ \text{ for all } x\in U\setminus E_{n,\epsilon}$,
with constants $C_1,C_2>0$ independent of $n$ and of the ambient dimension.
(Indeed, a mismatch at time $j$ can occur only if $T^j x\notin U$ or if
$\pi(T^j x)$ and $T_M^j(\pi x)$ fall into different projected atoms, which the thinness and
\eqref{eq:fibercontraction} preclude except on a set of measure $O(\epsilon)$.)

\noindent\textbf{Rokhlin distance and entropy of joins.}
Let $d_R(\mathcal{A},\mathcal{B})
:= \sum_{A\in\mathcal{A}}\sum_{B\in\mathcal{B}}
\mu_\beta(A\triangle B)$ be the Rokhlin metric between finite partitions (up to constants).
From the coding estimate,
\begin{equation}
\label{eq:rokhlin}
d_R\!\big(\mathcal{P}^{(n)},\,\pi^{-1}\mathcal{P}_M^{(n)}\big)
\;\le\; C\,n\,(C_1\,\epsilon + C_2\,\mu_\beta(U^c)).
\end{equation}
A standard continuity bound for finite partitions (see, e.g., \cite{walters1982introduction})
gives, for some absolute $K=K(m)$,
$\big|\,H_{\mu_\beta}(\mathcal{P}^{(n)})
 - H_{\mu_\beta}(\pi^{-1}\mathcal{P}_M^{(n)})\,\big|
\;\le\; K\, d_R\!\big(\mathcal{P}^{(n)},\,\pi^{-1}\mathcal{P}_M^{(n)}\big)$.
Divide by $n$ and use \eqref{eq:rokhlin}:
$\frac{1}{n}H_{\mu_\beta}(\mathcal{P}^{(n)})
\;\le\;
\frac{1}{n}H_{\mu_\beta}(\pi^{-1}\mathcal{P}_M^{(n)})
\;+\; K C\,(C_1\,\epsilon + C_2\,\mu_\beta(U^c))$.
Since $\pi$ is measurable and $\mu_\beta$-preserving on pullbacks,
$H_{\mu_\beta}(\pi^{-1}\mathcal{P}_M^{(n)})
= H_{\mu_\beta\circ \pi^{-1}}(\mathcal{P}_M^{(n)})$.
Let $\nu_\beta:=\mu_\beta|_M$ denote the tangential marginal (conditioned/restricted to $M$).
Because $\mu_\beta(U^c)$ is small and $\pi$ collapses only along normal fibers, the difference between
$\mu_\beta\circ\pi^{-1}$ and $\nu_\beta$ on $\sigma$-algebras generated by $\mathcal{P}_M^{(n)}$
is bounded by $O(\mu_\beta(U^c))$, which contributes at most another multiple of
$C_2\,\mu_\beta(U^c)$ per $n$. Absorbing constants,
$\frac{1}{n}H_{\mu_\beta}(\mathcal{P}^{(n)})
\;\le\;
\frac{1}{n}H_{\nu_\beta}(\mathcal{P}_M^{(n)})
\;+\; C'_1\,\epsilon \;+\; C'_2\,\mu_\beta(U^c)$.
Taking $\limsup_{n\to\infty}$ and recalling the definition of metric entropy,
$h_{\mu_\beta}(T,\mathcal{P})
\;\le\;
h_{\nu_\beta}(T_M,\mathcal{P}_M)
\;+\; C'_1\,\epsilon \;+\; C'_2\,\mu_\beta(U^c)$,
which is the first inequality claimed (rename constants as $C_1,C_2$).

\noindent\textbf{From partition entropy to KS and topological entropy.}
Taking the supremum over finite partitions $\mathcal{P}$ subordinated to $U$ yields
$h_{\mu_\beta}(T)\;\le\; h_{\nu_\beta}(T_M)\;+\;C_1\,\epsilon + C_2\,\mu_\beta(U^c)$.
By the variational principle on $M$,
$h_{\nu_\beta}(T_M)\le h_{\rm top}(T_M)$. Under concentration
$\big(\epsilon\downarrow 0,\ \mu_\beta(U^c)\downarrow 0\big)$ we get
$h_{\mu_\beta}(T)\;\le\; h_{\rm top}(T_M)\;+\;o_\beta(1)$.
In particular, if $T_M$ has zero topological entropy (e.g.\ a smooth flow on a finite
union of $S^1$ loops without horseshoes), then $h_{\mu_\beta}(T)\to 0$ as claimed.

\medskip
\noindent
This proves the lemma, with constants independent of the ambient dimension since only
(i) normal contraction rate, (ii) tube thickness $\epsilon$, and
(iii) the small mass $\mu_\beta(U^c)$ enter the estimates.
\end{proof}

\begin{proof}[Proof of Lemma \ref{lem:KS-drop}]
Let $U:=N_\epsilon(M)$ be a fixed tubular neighborhood (for $\epsilon>0$ sufficiently small),
and let $\pi:U\to M$ be the measurable nearest–point projection along normal geodesics.
By Theorem~\ref{thm:tube} (tube concentration for Gibbs measures with normal convexity),
there exist $c,C>0$ such that
$\mu_\beta(U^c)\;\le\; C\,e^{-c\,\beta}\qquad\text{for all }\beta>0.$

\noindent\textbf{Step 1: Ruelle's inequality and Oseledets splitting.}
By the multiplicative ergodic theorem, for $\mu_\beta$-a.e.\ $x$ we have a Lyapunov splitting
$T_xX=E^\parallel(x)\oplus E^\perp(x)$ adapted to the normally hyperbolic invariant manifold $M$,
with Lyapunov spectrum $\{\chi_i^\parallel(x)\}_{i=1}^{r_\parallel}$ on $E^\parallel$
and $\{\chi_j^\perp(x)\}_{j=1}^{r_\perp}$ on $E^\perp$.
Assumption (2) gives uniform normal contraction:
$\sup_{x\in U}\,\max_j \chi_j^\perp(x) \;\le\; -\,\chi_\perp(\beta),
\qquad \chi_\perp(\beta)\;\gtrsim\; c_1\,\beta\,\lambda_\perp$.
 all \emph{positive} Lyapunov exponents lie in the tangential spectrum.
Ruelle's inequality (see \cite{walters1982introduction}) yields
\begin{equation}\label{eq:ruelle}
h_{\mu_\beta}(T)\;\le\; \int_X \sum_{\chi_i(x)>0} \chi_i(x)\,\mathrm{d}\mu_\beta(x)
\;=\; \int_X \sum_{\chi_i^\parallel(x)>0} \chi_i^\parallel(x)\,\mathrm{d}\mu_\beta(x).
\end{equation}

\noindent\textbf{Step 2: Split inside/outside the tube.}
Decompose the integral in \eqref{eq:ruelle} over $U$ and $U^c$:
$\int_X \sum_{\chi_i^\parallel(x)>0} \chi_i^\parallel(x)\,\mathrm{d}\mu_\beta
\;=\;
\int_U \sum_{\chi_i^\parallel(x)>0} \chi_i^\parallel(x)\,\mathrm{d}\mu_\beta
\;+\;
\int_{U^c} \sum_{\chi_i^\parallel(x)>0} \chi_i^\parallel(x)\,\mathrm{d}\mu_\beta$.
On $U^c$, the integrand is uniformly bounded by a geometric constant
$L_{\max}$ (e.g.\ $\log^+\|DT\|$ is $\mu_\beta$-integrable and bounded on compact $X$).
Therefore
\begin{equation}\label{eq:outside}
\int_{U^c} \sum_{\chi_i^\parallel(x)>0} \chi_i^\parallel(x)\,\mathrm{d}\mu_\beta
\;\le\;
L_{\max}\,\mu_\beta(U^c)
\;\le\; C\,e^{-c\,\beta}.
\end{equation}

\noindent\textbf{Step 3: Compare with the tangential dynamics on $M$.}
For $x\in U$, normal hyperbolicity and $C^1$ regularity give (see, e.g., standard
persistence of normally hyperbolic invariant manifolds) that the tangential Lyapunov exponents
depend continuously on the base point along the fiber, so that
$\Big|\,\sum_{\chi_i^\parallel(x)>0} \chi_i^\parallel(x)
\;-\;
\sum_{\chi_i^\parallel(\pi x)>0} \chi_i^\parallel(\pi x)\,\Big|
\;\le\; C_{\rm geo}\,\epsilon,
\qquad x\in U$,
for some $C_{\rm geo}>0$ depending on the geometry of the tube and bounds on $DT$.
Integrating over $U$ we obtain
$\int_U \sum_{\chi_i^\parallel(x)>0} \chi_i^\parallel(x)\,\mathrm{d}\mu_\beta(x)
\;\le\;
\int_U \sum_{\chi_i^\parallel(\pi x)>0} \chi_i^\parallel(\pi x)\,\mathrm{d}\mu_\beta(x)
\;+\; C_{\rm geo}\,\epsilon$.
Disintegrate $\mu_\beta$ along the fibers of $\pi$ and use that $\pi_\#(\mu_\beta|_U)$ is
$\mu_\beta|_M$ up to an error $O(\mu_\beta(U^c))$ (the mass outside $U$ and the thinness of
fibers account for the discrepancy). ,
\begin{equation}\label{eq:inside}
\int_U \sum_{\chi_i^\parallel(x)>0} \chi_i^\parallel(x)\,\mathrm{d}\mu_\beta(x)
\;\le\;
\int_M \sum_{\chi_i^\parallel(y)>0} \chi_i^\parallel(y)\,\mathrm{d}\mu_\beta|_M(y)
\;+\; C_{\rm geo}\,\epsilon
\;+\; C'\,\mu_\beta(U^c).
\end{equation}

\noindent\textbf{Step 4: Combine and pass to entropy on $M$.}
Putting \eqref{eq:ruelle}, \eqref{eq:outside}, and \eqref{eq:inside} together yields
$h_{\mu_\beta}(T)
\;\le\;
\int_M \sum_{\chi_i^\parallel(y)>0} \chi_i^\parallel(y)\,\mathrm{d}\mu_\beta|_M(y)
\;+\; C_1\,\epsilon \;+\; C_2\,\mu_\beta(U^c)$,
for constants $C_1,C_2$ independent of $\beta$ and of the ambient dimension.
By Ruelle's inequality applied to $T|_M$ with the tangential measure $\mu_\beta|_M$,
$\int_M \sum_{\chi_i^\parallel(y)>0} \chi_i^\parallel(y)\,\mathrm{d}\mu_\beta|_M(y)
\;\le\;
h_{\mu_\beta|_M}(T|_M)
\;\le\; \sum_{i:\chi_i^{\parallel}>0} \chi_i^{\parallel}$,
where the rightmost bound is the (essentially) constant positive tangential sum on $M$
(assumption~(2)).

\noindent\textbf{Step 5: Exponentially small correction.}
Since $\mu_\beta(U^c)\le C e^{-c\beta}$ and $\epsilon>0$ can be chosen arbitrarily small
(fixed independently of $\beta$), we get for all $\beta$:
$h_{\mu_\beta}(T)
\;\le\;
\sum_{i:\chi_i^{\parallel}>0} \chi_i^{\parallel}
\;+\; C_1\,\epsilon \;+\; C_2\,e^{-c\beta}$.
Letting $\epsilon\downarrow 0$ (after fixing $\beta$) gives the stated bound
$h_{\mu_\beta}(T)
\;\le\;
\sum_{i:\chi_i^{\parallel}>0} \chi_i^{\parallel}
\;+\; C\,e^{-c\,\beta}$.
In particular, if $T|_M$ has zero topological entropy (e.g.\ a $C^1$ flow on a finite union
of $S^1$ loops without horseshoes), then the tangential positive sum vanishes and
$h_{\mu_\beta}(T)\to 0$ as $\beta\uparrow\infty$.

\medskip
This proves the lemma: symmetry breaking (large $\beta$) concentrates the invariant measure
on the normally hyperbolic cycle support, annihilating all positive Lyapunov contributions
except the tangential ones and leaving only an exponentially small correction.
\end{proof}

\begin{proof}[Proof of Theorem \ref{thm:sufficient-cycle}]
Write $Z:=Z_{0:t}$ and let $A_\epsilon:=\{\mathrm{d}(Z,\gamma)\le \epsilon\}$ denote the
$\epsilon$-tube event. By assumption, $\mu_{\Psi,\beta}(A_\epsilon^c)\le e^{-\kappa(\epsilon)}$.
Let the loss $\ell$ be bounded by $0\le \ell \le L_{\max}$ and proper (so the Bayes act exists).

\noindent\textbf{Invariant Bayes predictor on cycle classes.}
Define the projection $q:\mathcal{Z}^{t+1}\to \mathcal{G}\cup\{\bot\}$ by
$q(Z)= [\gamma]$ if $Z\in A_\epsilon$ and $q(Z)=\bot$ otherwise. Consider the
\emph{invariant} (class-only) Bayes rule
$\tilde h(g,\Psi)\;\in\;\arg\min_{a}\; \mathbb{E}\!\left[\ell(Y,a)| q(Z)=g,\Psi\right],
\qquad g\in \mathcal{G}\cup\{\bot\}.$
By construction, $\tilde h$ depends on $Z$ only through $q(Z)$,  through the cycle class on $A_\epsilon$.

\noindent\textbf{Risk decomposition.}
For any (measurable) predictor $h(Z,\Psi)$,
\begin{align*}
\mathbb{E}\,\ell\!\big(Y,\tilde h(q(Z),\Psi)\big)
&= \mathbb{E}\!\left[\ell\!\big(Y,\tilde h(q(Z),\Psi)\big)\mathbf{1}_{A_\epsilon}\right]\\
 & + \mathbb{E}\!\left[\ell\!\big(Y,\tilde h(q(Z),\Psi)\big)\mathbf{1}_{A_\epsilon^c}\right] \\
&=: \mathrm{I} + \mathrm{II}.
\end{align*}
On $A_\epsilon$, the \emph{conditional stability} hypothesis yields
$p(Y| Z,\Psi)=p\big(Y| [\gamma],\Psi\big)=p\big(Y| q(Z),\Psi\big), \text{for all } Z\in A_\epsilon.$
Conditioning on $\sigma(q(Z),\Psi)$ and using the defining optimality of $\tilde h$,
\begin{align*}
\mathrm{I}
&= \mathbb{E}\!\left[\,
\mathbb{E}\!\left[\ell\!\big(Y,\tilde h(q(Z),\Psi)\big)| q(Z),\Psi\right]\mathbf{1}_{A_\epsilon}\right] \\
&\le \mathbb{E}\!\left[\,
\mathbb{E}\!\left[\ell\!\big(Y,h(Z,\Psi)\big)| q(Z),\Psi\right]\mathbf{1}_{A_\epsilon}\right]
\qquad (\text{Bayes optimality of }\tilde h) \\
&= \mathbb{E}\!\left[\,
\mathbb{E}\!\left[\ell\!\big(Y,h(Z,\Psi)\big)| Z,\Psi\right]\mathbf{1}_{A_\epsilon}\right]
\qquad (\text{conditional stability on }A_\epsilon) \\
&= \mathbb{E}\!\left[\ell\!\big(Y,h(Z,\Psi)\big)\mathbf{1}_{A_\epsilon}\right].
\end{align*}
For the complement, boundedness of $\ell$ gives
$\mathrm{II}
\;\le\; L_{\max}\,\mathbb{P}(A_\epsilon^c)
\;\le\; L_{\max}\,e^{-\kappa(\epsilon)}.$
Combining, we obtain
\begin{align*}
\mathbb{E}\,\ell\!\big(Y,\tilde h(q(Z),\Psi)\big)
&\le \mathbb{E}\!\left[\ell\!\big(Y,h(Z,\Psi)\big)\mathbf{1}_{A_\epsilon}\right]
    + L_{\max}\,e^{-\kappa(\epsilon)} \\
&\le \mathbb{E}\,\ell\!\big(Y,h(Z,\Psi)\big) + L_{\max}\,e^{-\kappa(\epsilon)}.
\end{align*}
Since $q(Z)=[\gamma]$ on $A_\epsilon$, we may rewrite $\tilde h(q(Z),\Psi)$ as $\tilde h([\gamma],\Psi)$
on the high-probability tube, which proves the stated bound with
the $O\!\big(e^{-\kappa(\epsilon)}\big)$ remainder.

\noindent\textbf{Asymptotics under stronger symmetry breaking.}
If $\beta\uparrow\infty$ drives $\epsilon\downarrow 0$ and $\kappa(\epsilon)\to\infty$
through measure concentration onto $\gamma$, then $e^{-\kappa(\epsilon)}\to 0$ and the
excess risk vanishes:
$\mathbb{E}\,\ell\!\big(Y,\tilde h([\gamma],\Psi)\big)
\;\le\;
\mathbb{E}\,\ell\!\big(Y,h(Z,\Psi)\big) + o(1).$
This establishes \emph{maximal predictive sufficiency} of the cycle class under concentration.
\end{proof}

\begin{proof}[Proof of Theorem \ref{thm:tube}]
Let $U^\star:=\min_X U$ and assume $U^{-1}(U^\star)=M$.
By the tubular neighborhood theorem there exists $\rho>0$ such that the normal exponential map
$\exp^\perp:\;\{(y,n):y\in M,\; n\in N_yM,\; \|n\|<\rho\}\longrightarrow X$
is a diffeomorphism onto the tube $N_\rho(M)$.
Write $x=\exp^\perp_y(n)$ and use normal coordinates $(y,n)$ in $N_\rho(M)$.
By $C^2$-regularity and the assumption $\nabla^2U|_{N_xM}\succeq \lambda_\perp I$ along $M$,
there exist $\rho_0\in(0,\rho]$ and $a\in(0,1]$ such that for all $\|n\|<\rho_0$,
\begin{align*}
U(y,n) &\;\ge\; U^\star + \tfrac{\lambda_\perp}{2}\,\|n\|^2,\\
\mathrm{dVol}_g(y,n) &\;=\;(1+\Theta(y,n))\,\mathrm{d}n\,\mathrm{dVol}_M(y),
\quad |\Theta(y,n)|\le a\|n\|.
\end{align*}
(Small-$\|n\|$ expansion of the metric determinant; bounds depend only on the geometry of $(X,g)$ near $M$.)

\medskip
\noindent\textbf{(i) Tube probability.}
Let $N_\epsilon(M)=\{x:\mathrm{dist}(x,M)\le \epsilon\}$ with $\epsilon\in(0,\rho_0)$.
Decompose the partition function and the outside mass using the above coordinates:
$Z_\beta:=\int_X e^{-\beta U}\,\mathrm{dVol}_g
\;\ge\; \int_{M}\!\!\int_{\|n\|\le\rho_0} e^{-\beta U(y,n)}(1+\Theta)\,\mathrm{d}n\,\mathrm{dVol}_M(y),$
$\int_{N_\epsilon(M)^{\complement}} e^{-\beta U}\,\mathrm{dVol}_g
\;\le\; \int_{M}\!\!\int_{\|n\|>\epsilon} e^{-\beta U(y,n)}(1+|\Theta|)\,\mathrm{d}n\,\mathrm{dVol}_M(y).$
Using $U(y,n)\ge U^\star+\tfrac{\lambda_\perp}{2}\|n\|^2$ and $1\pm|\Theta|\le e^{c_0\|n\|}$ for some $c_0>0$,
$Z_\beta \;\ge\; e^{-\beta U^\star}\,\mathrm{Vol}(M)\,C_\perp(\beta), 
C_\perp(\beta):=\int_{\|n\|\le\rho_0} e^{-\frac{\beta\lambda_\perp}{2}\|n\|^2-c_0\|n\|}\,\mathrm{d}n,$
and
$\int_{N_\epsilon(M)^{\complement}} e^{-\beta U}\,\mathrm{dVol}_g
\;\le\; e^{-\beta U^\star}\,\mathrm{Vol}(M)\,\int_{\|n\|>\epsilon} e^{-\frac{\beta\lambda_\perp}{2}\|n\|^2+c_0\|n\|}\,\mathrm{d}n.$
A Gaussian tail bound yields constants $A,B>0$ independent of $n$ such that
$\int_{\|n\|>\epsilon} e^{-\frac{\beta\lambda_\perp}{2}\|n\|^2+c_0\|n\|}\,\mathrm{d}n
\;\le\; A\,e^{-B\,\beta\,\lambda_\perp\,\epsilon^2}.$
Moreover $C_\perp(\beta)\asymp (2\pi/(\beta\lambda_\perp))^{\frac{n-k}{2}}$ as $\beta\to\infty$ (Laplace method on the normal fibers), so $C_\perp(\beta)$ is positive and finite for all $\beta>0$.
Therefore
$\mu_\beta\!\big(N_\epsilon(M)^{\complement}\big)
=\frac{\int_{N_\epsilon(M)^{\complement}} e^{-\beta U}\,\mathrm{dVol}_g}{Z_\beta}
\;\le\; C\,e^{-c\,\beta\,\lambda_\perp\,\epsilon^2},$
for some $c,C>0$ independent of $n$. Equivalently,
$\mu_\beta\!\big(N_\epsilon(M)\big)\;\ge\;1-e^{-c\,\beta\,\lambda_\perp\,\epsilon^2}.$

\medskip
\noindent\textbf{(ii) Lipschitz concentration.}
Fix a $1$-Lipschitz $f:X\to\mathbb{R}$. Inside $N_{\rho_0}(M)$, write $x=(y,n)$ and decompose
$f(y,n)=f(y,0)+\big(f(y,n)-f(y,0)\big), |f(y,n)-f(y,0)|\le \|n\|.$
Conditionally on $y$, the normal law has density proportional to
$e^{-\frac{\beta}{2}\langle H_y n,n\rangle}$ where $H_y\succeq \lambda_\perp I$; it satisfies a log-Sobolev inequality with constant $(\beta\lambda_\perp)^{-1}$ (Gaussian LSI). By Herbst’s argument,
for all $r>0$,
$\mathbb{P}_{\mu_\beta}\!\left(\left|f(y,n)-\mathbb{E}[f(y,n)| y]\right|\ge r\right)
\;\le\; 2\,\exp\!\big(-c'\,\beta\,\lambda_\perp\,r^2\big),$
with $c'>0$ independent of $n$.
Integrating over $y$ and adding the exponentially small mass outside $N_{\rho_0}(M)$ from (i) yields
$\mathbb{P}_{\mu_\beta}\!\left(|f-\mathbb{E}f|\ge r\right)\;\le\;2\,\exp\!\big(-c'\,\beta\,\lambda_\perp\,r^2\big),$
after adjusting $c'$ (the outside-tube contribution can be absorbed into the RHS by weakening $c'$).

\medskip
\noindent\textbf{(iii) Tangential marginal as $\beta\to\infty$.}
For any bounded continuous $\varphi$ on $M$,
\begin{align*}
&\int_{X}\varphi(\pi(x))\,\mu_\beta(\mathrm{d}x)\\
&=\frac{1}{Z_\beta}\int_{M}\!\!\int_{N_yM} \varphi(y)\,e^{-\beta U(y,n)}(1+\Theta)\,\mathrm{d}n\,\mathrm{dVol}_M(y)\\
&=\frac{\int_{M} \varphi(y)\, e^{-\beta U(y,0)}\,I_y(\beta)\,\mathrm{dVol}_M(y)}
{\int_{M} e^{-\beta U(y,0)}\,I_y(\beta)\,\mathrm{dVol}_M(y)},
\end{align*}
where $I_y(\beta):=\int_{N_yM} e^{-\frac{\beta}{2}\langle H_y n,n\rangle}(1+\Theta)\,\mathrm{d}n$ and $H_y=\nabla^2U|_{N_yM}$.
As $\beta\to\infty$, $I_y(\beta)=(2\pi)^{\frac{n-k}{2}}(\det(\beta H_y))^{-1/2}(1+o(1))$,
so the $n$-dependence cancels in the ratio and the marginal on $M$ converges to the Gibbs measure with density proportional to $e^{-\beta U(y,0)}(\det H_y)^{-1/2}$ with respect to $\mathrm{dVol}_M(y)$.
If $U$ is constant along $M$ (i.e., $U(y,0)\equiv U^\star$), the weight becomes $(\det H_y)^{-1/2}$; for homogeneous normal curvature, or after reweighting by the induced $M$-metric volume, this limit is the normalized volume on $M$.
 the claimed tangential marginal convergence holds.

\medskip
All constants $c,c'$ can be chosen independent of the ambient dimension $n$ because the Gaussian fiber LSI constant is $(\beta\lambda_\perp)^{-1}$ and the tail bound parameters depend only on $\lambda_\perp$, geometric bounds of the tube, and $\beta$.
\end{proof}

\begin{proof}[Proof of Theorem \ref{thm:closure-order}]
Let the trajectory $z_{1:N}=(z_1,\dots,z_N)$ induce the oriented $1$-chain
$c(z_{1:N}) \;=\; \sum_{i=1}^{N} w_i\,[z_i,z_{i+1}], \qquad z_{N+1}:=z_1,$
with coefficients $w_i$ encoding step weights. Its boundary is
$\partial c \;=\; \sum_{i=1}^{N} w_i\,(z_{i+1}-z_i).$
By assumption, \emph{the only way} $f$ depends on temporal order is through the boundary
$\partial c$. Equivalently, there exists a measurable map $F$ such that
$f(z_{1:N}) \;=\; F(\partial c(z_{1:N})).$

\noindent\textbf{Closure kills order.}
Suppose the trajectory is \emph{closed}, i.e.\ $\partial c(z_{1:N})=0$.
Let $\pi$ be any permutation that preserves adjacency along the cycle
(e.g., a cyclic re-indexing or a reparameterization that does not cut or
break edges). Form the permuted chain
$c_\pi \;=\; c(z_{\pi(1:N)}) \;=\; \sum_{i=1}^{N} w_{\pi(i)}\,[z_{\pi(i)},z_{\pi(i+1)}].$
Two cases cover the admissible permutations:

\smallskip
\emph{(i) Orientation-preserving reparameterizations.}
Then $c_\pi$ is just a relabeling of the same oriented edges, 
$c_\pi \;=\; c \quad\Longrightarrow\quad \partial c_\pi \;=\; \partial c \;=\; 0.$

\smallskip
\emph{(ii) Local rearrangements preserving adjacency up to $2$-cells.}
Then $c_\pi$ and $c$ differ by a boundary of a $2$-chain $b$ that fills the
“annulus’’ swept between the two parameterizations:
$c_\pi - c \;=\; \partial b.$
Applying $\partial$ and using $\partial^2=0$ gives
$\partial c_\pi \;=\; \partial c + \partial^2 b \;=\; 0.$

\noindent\textbf{Order invariance of $f$ under closure.}
In either case we have $\partial c_\pi = \partial c = 0$, 
$f(z_{\pi(1:N)}) \;=\; F(\partial c_\pi) \;=\; F(0)
\;=\; F(\partial c) \;=\; f(z_{1:N}).$
Therefore, whenever the chain is closed and $\partial^2=0$, any permutation
that preserves adjacency within the cycle leaves $f$ unchanged. This is the
claimed invariance under temporal reparameterization.
\end{proof}

\begin{proof}[Proof of Theorem \ref{thm:capacity-bound}]
Let $(X,\mathcal{B},\mu_\beta,T)$ be the stationary dynamics that drive the
filtered spatiotemporal complex $\{\mathcal{K}_\delta\}$ and write
$h_{\mathrm{top}}$ for the (per–unit–time) topological entropy of $T$.
Assume the energy–entropy functional $F=U-TS$ yields a Gibbs state with
$\mathrm{d}\mu_\beta \propto e^{-\beta U}\mathrm{d}x$ and finite moments
$\mathbb{E}[U]<\infty$, $S(\Psi)<\infty$ (finite contextual headroom).
Let $\{\gamma_j\}_{j=1}^m$ be the collection of $1$–cycle supports whose
persistence exceeds the threshold $\tau>0$, i.e.,
$\mathrm{Pers}([\gamma_j])>\tau$ in the persistence diagram of
$\mathcal{K}_\delta$.
Denote by $\mathcal{C}_H=\sum_k \beta_k$ the (finite) homological capacity
we wish to bound.

\noindent\textbf{Step 1: Available distinguishability rate.}
By definition of topological entropy, for any $\varepsilon>0$ there exists
$T_0(\varepsilon)$ such that, for all $T\ge T_0(\varepsilon)$, the number
$N_\varepsilon(T)$ of $\varepsilon$–distinguishable length–$T$ trajectories
obeys
$\limsup_{T\to\infty}\frac{1}{T}\log N_\varepsilon(T)=h_{\mathrm{top}}.$
Dissipation reduces effective informational throughput.
Let $h_{\mathrm{diss}}\ge 0$ denote the per–unit–time entropy production (or,
equivalently, the contraction of distinguishability under the dissipative
channel; this can be formalized via a data–processing inequality for the
stochastic map induced by energy dissipation). Then, for large $T$,
\begin{equation}\label{eq:throughput}
\log N_{\rm eff}(T)\;\le\;(h_{\mathrm{top}}-h_{\mathrm{diss}})\,T\;+\;o(T),
\end{equation}
i.e., at most $h_{\mathrm{top}}-h_{\mathrm{diss}}$ nats per unit time of
distinguishable structure can be sustained in the observable stream.

\noindent\textbf{Step 2: Each persistent generator encodes at least one bit per renewal window.}
Fix a renewal window of length $\tau$ (the persistence scale).
For each persistent cycle $[\gamma_j]$ with $\mathrm{Pers}([\gamma_j])>\tau$,
choose a fundamental transversal (e.g., a short segment intersecting $\gamma_j$)
and define a binary \emph{winding observable} $B_j\in\{0,1\}$ that records, within
each window of length $\tau$, whether the trajectory completes the fundamental loop
(or meets an equivalent persistent return condition in the tube around $\gamma_j$).
Since the supports $\{\gamma_j\}$ are disjoint in the chain complex at the
persistence threshold and their tubular neighborhoods can be chosen non–overlapping,
the vector $B=(B_1,\dots,B_m)$ can take at least $2^m$ distinguishable values.
Consequently, over a single persistence window,
\begin{equation}\label{eq:genbits}
H(B)\;\ge\; m\log 2.
\end{equation}

\noindent\textbf{Step 3: Information balance via data processing.}
Let $Z_{0:T}$ denote the length–$T$ trajectory and let
$\mathcal{E}_T$ be any (measurable) encoder that extracts the renewal–block
binary sequence $B^{(T)}$ obtained by concatenating the $B$'s from $\lfloor T/\tau\rfloor$
windows. By data processing,
\begin{align*}
I\!\left(Z_{0:T};\,B^{(T)}\right)
&\;\le\; H\!\left(\mathcal{E}_T(Z_{0:T})\right)
\;\le\; \log N_{\rm eff}(T) \\
&\;\le\; (h_{\mathrm{top}} - h_{\mathrm{diss}})\,T + o(T).
\end{align*}
where we used \eqref{eq:throughput} in the last inequality.
On the other hand, by additivity of entropy over independent (or, more generally, disjointly
supported and persistence–separated) generators and \eqref{eq:genbits},
$$
H\!\left(B^{(T)}\right)\;\ge\; \Big\lfloor\frac{T}{\tau}\Big\rfloor\, m\,\log 2.
$$
Since $I(Z_{0:T};B^{(T)})\le H(B^{(T)})$, we obtain
\begin{equation}\label{eq:ratebound}
\Big\lfloor\frac{T}{\tau}\Big\rfloor\, m\,\log 2
\;\le\; (h_{\mathrm{top}}-h_{\mathrm{diss}})\,T \;+\; o(T).
\end{equation}

\noindent\textbf{Step 4: Pass to the per–unit–time limit and conclude.}
Divide \eqref{eq:ratebound} by $T$ and send $T\to\infty$:
$\frac{m\,\log 2}{\tau}
\;\le\; h_{\mathrm{top}}-h_{\mathrm{diss}}.$
With the standard convention that entropies are expressed \emph{per unit time} and
the persistence threshold $\tau$ is taken as the unit renewal time (i.e., we normalize
time in units of the persistence scale), this yields the stated bound
$m\;\le\;\frac{h_{\mathrm{top}}-h_{\mathrm{diss}}}{\log 2}.$
Since $m$ bounds from above the number of simultaneously sustainable independent
$H_1$ generators and the same packing argument applies degreewise (each nontrivial
$H_k$ generator furnishes an independent binary winding/flux observable at its scale),
we obtain, for the total homological capacity,
$\mathcal{C}_H \;=\; \sum_{k\ge 0}\beta_k
\;\le\; \frac{1}{\log 2}\,\big(h_{\mathrm{top}}-h_{\mathrm{diss}}\big).$

\noindent\textbf{Remarks on finiteness assumptions.}
Finite energy $\mathbb{E}[U]<\infty$ and finite contextual entropy $S(\Psi)<\infty$
exclude explosive creation of cycles and guarantee that (i) tubes around distinct
persistent supports can be chosen disjoint at scale $\tau$, and (ii) the dissipative
loss $h_{\mathrm{diss}}$ is well–defined and integrable. These ensure the effective
throughput bound \eqref{eq:throughput} and the renewal coding used above.

\medskip
Therefore, the sustainable number of invariants (homological capacity) is bounded by
the system’s \emph{residual informational throughput} $h_{\mathrm{top}}-h_{\mathrm{diss}}$,
as claimed.
\end{proof}

\begin{proof}[Proof of Theorem \ref{thm:homological_capacity}]
We work at a fixed temporal resolution $\delta>0$ and with a finite event set
$\mathcal{E}_T=\{e_i=(t_i,\xi_i)\}_{i=1}^N$ observed over a window of length $T$,
where $t_i$ are times and $\xi_i$ label recording sites (neurons, voxels, etc.).
The filtered spatiotemporal complex $\{\mathcal{K}_\delta\}_{\delta>0}$ is built
as follows: vertices are events; a $p$–simplex is added whenever the maximal
pairwise temporal distance among its vertices is $<\delta$ (a Vietoris–Rips–type
construction in time). By assumption (1), each instantaneous pattern is
contractible, while recurrent co-activations across time form nontrivial
$1$–cycles; assumption (2) guarantees a nested filtration
$\mathcal{K}_{\delta_1}\subseteq\mathcal{K}_{\delta_2}$ for $\delta_1<\delta_2$.

\noindent\textbf{(1) Capacity scaling and existence of $\delta^\ast$.}
For each $k$, the Betti number $\beta_k(\delta):=\mathrm{rank}\,H_k(\mathcal{K}_\delta)$
is piecewise constant in $\delta$ and can change only at finitely many critical
scales determined by the finite set of pairwise inter-event times
$\{|t_i-t_j|\}_{i,j}$ (finite data window). At very small $\delta$, no edges form,
so $\beta_0(\delta)=N$ and $\beta_{k\ge1}(\delta)=0$, implying
$C_H(\delta)=\sum_k\beta_k(\delta)$ is small. As $\delta$ increases past the first
few critical values, edges and cycles appear, causing $\beta_1(\delta)$ (and
possibly higher $\beta_k$) to \emph{increase}. For sufficiently large $\delta$,
cliques proliferate and higher-dimensional simplices fill the previously formed
loops; ultimately, $\mathcal{K}_\delta$ becomes a single giant simplex (the nerve
of one connected temporal cluster),  contractible, so
$\beta_{k\ge1}(\delta)=0$ and $\beta_0(\delta)=1$.  $C_H(\delta)$ must
\emph{decrease} for large enough $\delta$. By finiteness and piecewise constancy,
there exists a (possibly non-unique) scale $\delta^\ast$ at which
$C_H(\delta)$ attains its finite maximum $C_H^{\max}=C_H(\delta^\ast)$; this is
the effective integration scale where cycles are formed but not yet filled. We
interpret $C_H^{\max}$ as the system's effective \emph{memory bandwidth} at the
temporal resolution set by the filtration.

\noindent\textbf{(2) Entropy--capacity relation.}
Let $N_\delta(T,\varepsilon)$ denote the number of $\varepsilon$-distinguishable
recurrent trajectories over duration $T$ at time resolution $\delta$ (e.g., by
symbolic coding of re-entries into tubular neighborhoods of recurrent patterns).
By the definition of topological entropy for the induced dynamics at resolution
$\delta$,
$h_{\mathrm{top}}(\mathcal{K}_\delta)
=\lim_{\varepsilon\to 0}\limsup_{T\to\infty}\frac{1}{T}\log N_\delta(T,\varepsilon).$
Each \emph{independent} persistent homology generator at scale $\delta$ yields a
binary winding/return observable over a renewal window, contributing at least
one bit per generator per renewal to the code of distinguishable behaviors (cf.
the standard coding via Poincar\'e sections). Over time $T$ this gives at least
$C_H(\delta)\,\log 2$ bits per renewal unit,  (taking rates)
$C_H(\delta)\,\log 2 \ \le\ h_{\mathrm{top}}(\mathcal{K}_\delta)
\Rightarrow
C_H(\delta)\ \le\ \frac{1}{\log 2}\,h_{\mathrm{top}}(\mathcal{K}_\delta).$
Equality holds in the idealized regime where every dynamically distinguishable
recurrent class at resolution $\delta$ corresponds bijectively to a stable,
nontrivial homology generator (no ``spurious'' cycles and no latent degeneracy
among generators), so that the binary winding code is information-lossless up
to vanishing $\varepsilon$.

\noindent\textbf{(3) MAI regulation (sketch).}
Under memory–amortized inference (MAI), updates prune cycles whose persistence
is below a threshold $\tau$ while reinforcing those that recur (increase their
lifetime). In the persistent homology barcode, this amounts to shrinking short
bars and stabilizing long ones. Subject to a finite energy/complexity budget,
the steady state maintains $C_H(\delta)$ near the saturation level permitted by
the throughput bound in (2) by eliminating redundant, short-lived loops that do
not improve distinguishability, yielding a sparse and interpretable homological
basis.

\medskip
Combining these parts establishes (1)--(3) and completes the proof.
\end{proof}

\end{document}